\documentclass[sigconf]{acmart}

\usepackage{MnSymbol}
\usepackage{amsmath}
\usepackage{amsfonts}

\usepackage{amssymb}                        % 数学符号生成命令
\usepackage{textcomp}
\usepackage{stfloats}
\usepackage{url}
\usepackage{verbatim}
\def\BibTeX{{\rm B\kern-.05em{\sc i\kern-.025em b}\kern-.08em
    T\kern-.1667em\lower.7ex\hbox{E}\kern-.125emX}}
\usepackage{balance}

\usepackage{algorithm}
\usepackage{algorithmic}
\usepackage{listings}
\usepackage{hyperref}
\usepackage{mathtools}
\usepackage{multirow}
\usepackage{bm}

\usepackage{microtype}
\usepackage{float}
\usepackage{color, colortbl}
\usepackage{stfloats}
\usepackage{enumitem}
\usepackage{xspace}
\usepackage{subcaption}
\usepackage{xcolor}
\usepackage{dsfont}
\usepackage{array}
\usepackage{pifont}
\usepackage[T1]{fontenc}
\usepackage{fancyhdr}

\newtheorem{proposition}{Proposition}

\newtheorem{remark}{Remark}

\newtheorem{theorem}{Theorem}
\newtheorem{definition}{Definition}

\AtBeginDocument{%
  \providecommand\BibTeX{{%
    Bib\TeX}}}

\copyrightyear{2023} 
\acmYear{2023} 
\setcopyright{acmlicensed}\acmConference[MM '23]{Proceedings of the 31st
ACM International Conference on Multimedia}{October 29-November 3,
2023}{Ottawa, ON, Canada}
\acmBooktitle{Proceedings of the 31st ACM International Conference on
Multimedia (MM '23), October 29-November 3, 2023, Ottawa, ON, Canada}
\acmPrice{15.00}
\acmDOI{10.1145/3581783.3612225}
\acmISBN{979-8-4007-0108-5/23/10}

\begin{document}

\setlength{\parskip}{0pt plus0.5pt minus5.4pt}
\setlength{\parskip}{0.15pt plus1pt minus1.8pt}
%%
%% The "title" command has an optional parameter,
%% allowing the author to define a "short title" to be used in page headers.
\title{Generalized Universal Domain Adaptation with Generative Flow Networks}

%%
%% The "author" command and its associated commands are used to define
%% the authors and their affiliations.
%% Of note is the shared affiliation of the first two authors, and the
%% "authornote" and "authornotemark" commands
%% used to denote shared contribution to the research.
\author{Didi Zhu}
\authornote{Both authors contributed equally to this research. This work was completed while
Didi Zhu was a member of the Huawei Noah’s Ark Lab for advanced study.}
% \orcid{1234-5678-9012}
% \authornotemark[1]
\affiliation{%
  \institution{Zhejiang University}
  % \streetaddress{P.O. Box 1212}
  \city{Hangzhou}
  % \state{Ohio}
  \country{China}
  % \postcode{43017-6221}
}
\email{didi_zhu@zju.edu.cn}

\author{Yinchuan Li}
\authornotemark[1]
\affiliation{%
  \institution{Huawei Noah’s Ark Lab}
  \city{Beijing}
  \country{China}}
\email{liyinchuan@huawei.com}

\author{Yunfeng Shao}
\affiliation{%
  \institution{Huawei Noah’s Ark Lab}
  \city{Beijing}
  \country{China}}
\email{shaoyunfeng@huawei.com}

\author{Jianye Hao}
\affiliation{%
 \institution{Tianjin University}
 \institution{Huawei Noah’s Ark Lab}
 \city{Tianjin}
 \country{China}}
 \email{jianye.hao@tju.edu.cn}

\author{Fei Wu}
\author{Kun Kuang}

\affiliation{%
  \institution{Zhejiang University}
  \city{Hangzhou}
  \country{China}
}
\email{wufei@zju.edu.cn}
\email{kunkuang@zju.edu.cn}

% \author{Kun Kuang}
% \affiliation{%
%   \institution{Zhejiang University}
%   \city{Hangzhou}
%   \country{China}
% }
% \email{kunkuang@zju.edu.cn}

\author{Jun Xiao}
\author{Chao Wu}
\authornote{Corresponding Author}
\affiliation{%
  \institution{Zhejiang University}
  \city{Hangzhou}
  \country{China}
}
\email{junx@cs.zju.edu.cn}
\email{chao.wu@zju.edu.cn}

% \author{Chao Wu}
% \authornote{Corresponding Author}
% \affiliation{%
%   \institution{Zhejiang University}
%   \city{Hangzhou}
%   \country{China}
% }
% \email{chao.wu@zju.edu.cn}

%%
%% By default, the full list of authors will be used in the page
%% headers. Often, this list is too long, and will overlap
%% other information printed in the page headers. This command allows
%% the author to define a more concise list
%% of authors' names for this purpose.
\renewcommand{\shortauthors}{Didi Zhu et al.}

%%
%% The abstract is a short summary of the work to be presented in the
%% article.
\begin{abstract}
  We introduce a new problem in unsupervised domain adaptation, termed as Generalized Universal Domain Adaptation (GUDA), which aims to achieve precise prediction of all target labels including unknown categories.
  GUDA bridges the gap between label distribution shift-based and label space mismatch-based variants, essentially categorizing them as a unified problem, guiding to a comprehensive framework for thoroughly solving all the variants.
  The key challenge of GUDA is developing and identifying novel target categories while estimating the target label distribution.
  To address this problem, we take advantage of the powerful exploration capability of generative flow networks and propose an active domain adaptation algorithm named GFlowDA, which selects diverse samples with probabilities proportional to a reward function.
  To enhance the exploration capability and effectively perceive the target label distribution, we tailor the states and rewards, and introduce an efficient solution for parent exploration and state transition.
  We also propose a training paradigm for GUDA called Generalized Universal Adversarial Network (GUAN), which involves collaborative optimization between GUAN and GFlowNet.
  Theoretical analysis highlights the importance of exploration, and extensive experiments on benchmark datasets demonstrate the superiority of GFlowDA.
\end{abstract}

\begin{CCSXML}
  <ccs2012>
  <concept>
  <concept_id>10010147.10010178.10010224</concept_id>
  <concept_desc>Computing methodologies~Computer vision</concept_desc>
  <concept_significance>500</concept_significance>
  </concept>
  </ccs2012>
\end{CCSXML}
  
\ccsdesc[500]{Computing methodologies~Computer vision}

%%
%% Keywords. The author(s) should pick words that accurately describe
%% the work being presented. Separate the keywords with commas.
\keywords{domain adaptation, label shift, open set, active learning}
%% A "teaser" image appears between the author and affiliation
%% information and the body of the document, and typically spans the
%% page.
% \begin{teaserfigure}
%   \includegraphics[width=\textwidth]{sampleteaser}
%   \caption{Seattle Mariners at Spring Training, 2010.}
%   \Description{Enjoying the baseball game from the third-base
%   seats. Ichiro Suzuki preparing to bat.}
%   \label{fig:teaser}
% \end{teaserfigure}

% \received{20 February 2007}
% \received[revised]{12 March 2009}
% \received[accepted]{5 June 2009}

%%
%% This command processes the author and affiliation and title
%% information and builds the first part of the formatted document.
\maketitle

\section{Introduction}
\label{sec:intro}
% Deep neural networks have been widely employed in various visual tasks and made revolutionary advances~\cite{deng2009large,krizhevsky2012imagenet,simonyan2014very,he2016deep,zhangfairness,zhang2020federated}.
Deep neural networks have been widely employed in various visual tasks and made revolutionary advances~\cite{deng2009large,krizhevsky2012imagenet,simonyan2014very,he2016deep,zhang2020federated,zhang2022fairness,shen2020federated,liu2022distributionallyb,zhang2022federated,zhang2022dense,zhang2020knowledge,zhang2022domain}.
However, deep learning algorithms highly rely on massive labeled data, and models trained with limited labeled data do not generalize well on domains with different distributions. 
Unsupervised domain adaptation (UDA)~\cite{ben2010theory} has emerged as a promising solution to address this limitation by adapting models trained on a source domain to perform well on an unlabeled target domain. 
% is a transfer learning technology that aims to minimize the distribution discrepancy between the source and target domain. 
Recent literature~\cite{ganin2016domain,long2015learning,tzeng2017adversarial,wang2014flexible,duan2012domain,saenko2010adapting,kang2018deep,shen2021towards,liu2021heterogeneous,liu2021kernelized,yuan2023instrumental,yuan2023domain,zhang2022tree,zhang2023rotogbml} addresses the UDA problem under covariate assumption~\cite{zhang2013domain}, for which methods perform importance weighting or aim at aligning the marginal distributions. However, these methods increase the general loss on the target domain when facing label heterogeneity.

UDA variants with label heterogeneity can be categorized based on the variation of label distribution, label space, and label prediction space.
The \textbf{\emph{label distribution}} can be split into conditions 
\ding{172} $P_s(Y) \neq P_t(Y)$ and \ding{173} $P_s(X|Y) \neq P_t(X|Y)$. 
$P_{\{s,t\}}(Y)$ and $P_{\{s,t\}}(X|$ $Y)$ indicate the margin label distribution and the class-conditional distribution, respectively.
The label shift (LS) problem~\cite{zhang2013domain} corresponds to \ding{172}, while the generalized label shift (GLS) problem~\cite{zhang2013domain} corresponds to \ding{172} $\wedge$ \ding{173}.
The \textbf{\emph{label space}} can be split into conditions: \ding{174} $\overline{\mathcal{Y}}_s  \neq \emptyset ~\&~ \text{unknown}$ and  
\ding{175}
$\overline{\mathcal{Y}}_t  \neq \emptyset ~\&~ \text{unknown}$ with  $\overline{\mathcal{Y}}_{\{s,t\}}$ denoting the private label space.
The \textbf{\emph{label prediction space}}, denoted as $|\widehat{\mathcal{Y}}|$, includes two conditions: \ding{176} $|\widehat{\mathcal{Y}}| = k+1$, where all target private labels are considered as "unknown"; and \ding{177} $|\widehat{\mathcal{Y}}| = k+n$, where $k$ and $n$ denote the size of  the common label set and the target private label set, respectively. 
Partial domain adaptation (PDA)~\cite{zhang2018importance} assumes private classes only exist in the source domain (\ding{174}), while open set domain adaptation (OSDA)~\cite{saito2018open} assumes they only exist in the target domain (\ding{175} $\wedge$ \ding{176}). Universal domain adaptation (UniDA)~\cite{you2019universal} is a more general case with no knowledge about the label space relationship, summarized as \ding{174} $\wedge$ \ding{175} $\wedge$ \ding{176}.

\begin{figure*}[tbp]
    \begin{center}
    \includegraphics[width=6.8in]{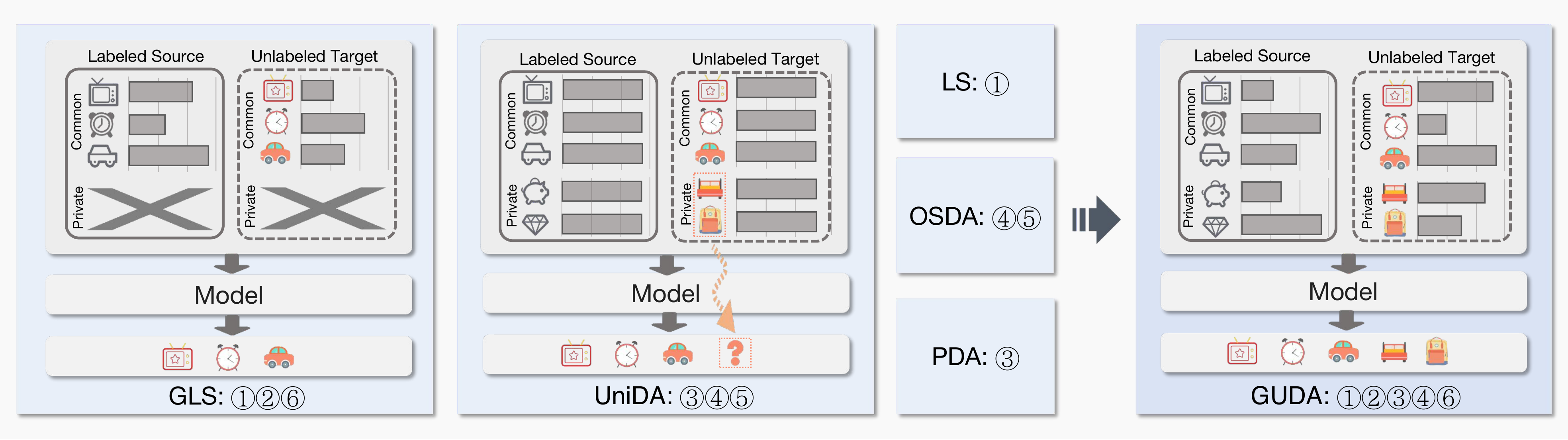}
    \end{center}
    \vspace{-0.2cm}
    \caption{Illustration of GUDA and some subproblems of GUDA. GUDA contains most domain adaptation tasks from the perspective of label distribution, label space and label prediction space.}
    \label{fig1}
    \vspace{-0.25cm}
  \end{figure*}

Label distribution based variants mainly focus on estimating label ratio to reweight source samples, whereas label space based variants aim to design heuristic rules to determine whether the samples belong to the private label set. 
However, methods for these two types of variants are incapable of resolving one another.
In practice, it is difficult to identify which variant is being encountered, and real-world scenarios may involve a combination of these variants, making it challenging to apply any single method. 
Furthermore, the label space based variants can only identify unseen categories as "unknown", limiting their ability to achieve fine-grained recognition. 
These dilemmas highlight the necessity for a unified framework to address both label distribution and label space variants, while also enabling precise prediction of unknown categories.

% Label Exploration
In this paper, we propose a new problem \textbf{\emph{Generalized Universal Domain Adaptation} (GUDA)}  to take a unified view of these above problems, as illustrated in Fig.~\ref{fig1}, which can be represented as \ding{172} $\wedge$ \ding{173} $\wedge$ \ding{174} $\wedge$ \ding{175} $\wedge$ \ding{177}. 
The key challenge of GUDA is {developing and identifying novel categories} that exist in the target domain while {estimating the overall label distribution} for subsequent  feature alignment.
% The goal of GUDA is to explore the entire target label distribution and label space to better perform subsequent \textbf{weighted feature alignment} and \textbf{unknown class recognition}. 
To address GUDA, we take the benefit of the powerful exploration capability of Generative Flow Networks (GFlowNets)~\cite{bengio2021flow} and propose an active domain adaptation framework named \textbf{{GFlowDA}},  which could explore the overall target label distribution by annotating a subset of target data. 
Unlike the previous active domain adaptation (ADA) works~\cite{fu2021transferable, 
 prabhu2021active,xie2022active,ma2021active, hwang2022combating,xie2022learning,de2021discrepancy}, which either focus on designing metrics~\cite{ma2021active,fu2020learning,xie2022learning} that may have some bias or minimizing the feature distance while easily falling into the trap of sub-optimal solutions~\cite{hwang2022combating,de2021discrepancy,prabhu2021active,xie2022active}, our insight is to learn a generative policy that generates a distribution  with probabilities proportional to the distance of the original target distribution.
We consider the selected target subset as a compositional object and formulate the ADA problem as a distribution generative process by sequentially selecting a target sample through GFlowNets. 
To facilitate GFlowNets to better perceive the target label distribution, we customize the states and rewards, and introduce an efficent parent exploration and state transition approach. Finally, we propose a weighted adaptive model named Generalized Universal Adversarial Network (GUAN), which enables efficient domain alignment through a reciprocal relationship between GFlowNets and GUAN.

\textbf{Main Contribution:} (1) We introduce GUDA, which covers most UDA variants with label heterogeneity and aims to recognize all target classes including unknown classes.
(2) To address this challenge, we propose GFlowDA to select and annotate target samples to estimate the overall target label distribution.
(3) We define the design paradigm for states and rewards, and introduce an efficient solution for parent exploration and state transition in the GFlowDA training process. We also propose a new training paradigm called GUAN, which involves collaborative optimization between GUAN and GFlowNet.
(4) Theoretical analysis and extensive experiments show that the effectiveness of our GFlowDA.

\section{Related Works}
\label{sec:related_work}

\subsection{Unsupervised Domain Adaptation}

For LS and GLS problems in the literature of UDA, most works ~\cite{tachet2020domain,shui2021aggregating,le2021lamda,kirchmeyer2021mapping,rakotomamonjy2022optimal, zhao2019learning} seek to estimate the label ratio  $P_t(Y)/P_s(Y)$ to weight the source feature, which requires that the source label distribution $P_s(Y)$ cannot be zero. This underlying constraint limits the generalization of the methods for LS and GLS to OSDA~\cite{saito2018open,panareda2017open} and UniDA~\cite{you2019universal} scenarios due to the existence of target private labels. 
% \textcolor{blue}{LS and GLS need explore the label distribution but fail on OSDA and UniDA.}
% OSDA~\cite{saito2018open,panareda2017open} and UniDA~\cite{you2019universal} divide the target samples into $k+1$ classes. 
Numerous methods for OSDA and UniDA either utilize prediction uncertainty~\cite{saito2018maximum, you2019universal,fu2020learning,lifshitz2020sample,saito2021ovanet,tao2023imbalanced,ma2022attention}, or incorporate self-supervised learning techniques~\cite{baktashmotlagh2018learning, saito2020universal, li2021domain, zhu2023universal,tong2023quantitatively}. However, these methods often fail to explore fine-grained discriminative knowledge in the unknown set and do not take label distribution shift  into account within the common label space. Additionally, most PDA methods~\cite{cao2018partial1, zhang2018importance, liang2020balanced}  aim to weight source samples with heuristic criteria, which suffe from the same limitations as OSDA and UniDA. 
While recent methods such as OSLS~\cite{garg2022domain} and AUDA~\cite{ma2021active} attempt to address some of the above limitations, they may not be suitable for GUDA. Further research is needed to develop more effective methods that can overcome the challenges posed by GUDA.

\subsection{Generative Flow Networks}
GFlowNets~\cite{bengio2021gflownet} is a generative model which aims to solve the problem of generating diverse
candidates. It has been effective in various fields such as molecule generation~\cite{bengio2021flow,malkin2022trajectory}, discrete probabilistic modeling~\cite{zhang2022generative}, graph neural networks~\cite{li2023dag,li2023generative} and causal discovery~\cite{li2022gflowcausal}. Another research direction aims to address and extend the inherent assumptions of the original GFlowNet~\cite{li2023cflownets, li2022generative,li2023gflownets,wang2023regularized}.
Compared to reinforcement learning (RL) methods \cite{sutton2018reinforcement}, which focus on maximizing the expected return by generating a sequence of actions with the highest reward, GFlowNets offer the ability to explore diverse reward distributions by sampling trajectories with probabilities proportional to the expected rewards. This feature allows for more effective estimation of the target distribution.
%  making GFlowNets particularly well-suited for GUDA. Subsequent experiments will further analyze the differences between RL and GFlowNets.

\subsection{Active Domain Adaptation} 
The pioneering study ~\cite{rai2010domain} demonstrates how active learning (AL) and DA can collaborate to enhance AL in DA. 
Recent efforts ~\cite{su2020active,fu2021transferable,xie2022active} design some criterion by introducing advanced techniques like adversarial training, multiple discriminators and free energy model. 
% However, these criteria can not guarantee that the chosen samples can accurately reflect the whole target distribution. 
Besides, some parallel works~\cite{prabhu2021active,ma2021active} suggest using clustering to choose samples.
%  However, these works have a heavy computational burden and are difficult to apply to large scale datasets.
 In addition, distance-based works \cite{de2021discrepancy,hwang2022combating,xie2022learning,prabhu2021active,ma2021active,yuan2022label} choose samples based on their distance to the source or the target domain. 
% However, these works have a heavy computational burden and also increase computing complexity and easily fall into the trap of sub-optimal solutions.
Overall, existing works mainly rely on manually-designed criteria or distance, leading to overfitting to specific scenarios and easily falling into sub-optimal solutions. 
Unlike the existing ADA methods, we consider the query batch as a compositional object and formulate the ADA as a generative process. The generative policy network can automatically explore how to find the most informative samples in an essential way.

\section{Problem Formulation}
\label{sec:method}

% To improve readability,  we provide a summary of the important notations in Table \ref{tab:notation} that are mentioned later in this paper.

% To increase the readability of this paper, we provide a summary of the important notations used throughout this work in Table \ref{tab:notation}.

Denoting $\mathcal{X}$, $\mathcal{Y}$, $\mathcal{Z}$ as the input space, label space and latent space, respectively. 
Let $X$, $Y$ and $Z$ be the random variables of $\mathcal{X}$, $\mathcal{Y}$, $\mathcal{Z}$, and  $x$, $y$ and $z$ be their respective elements.
% We denote the random variables of $\mathcal{X}$ $\mathcal{Y}$, $\mathcal{Z}$ as $X$, $Y$ and $Z$, and their respective elements as $x$, $y$ and $z$.
Let $P_s$ and $P_t$ be the source distribution and target distribution. 
% Random variables of  $\mathcal{X}$, $\mathcal{Y}$, $\mathcal{Z}$  are noted as $X$, $Y$ and $Z$. Elements of  $\mathcal{X}$, $\mathcal{Y}$, $\mathcal{Z}$  are noted as $x$, $y$ and $z$.
%  Let $P_s$ and $P_t$ be the source distribution and target distribution and let $P_s$ and $P_t$ denote the corresponding probability density function. 
We are given a labeled source domain $\mathcal{D}_{s}= \{{x}_{i}, {y}_{i})\}_{i=1}^{m}$ and an unlabeled target domain $\mathcal{D}_{t}=  \{{x}_{i}\}_{i=1}^{n}$ are respectively sampled from $P_{s}$ and $P_{t}$, where $m$ and $n$ denote the numbers of source samples and target samples. Denote $\mathcal{Y}_{s}$ and $\mathcal{Y}_{t}$ as the label sets of the source and target domains, respectively.  Suppose the feature transformation function is $g: \mathcal{X} \rightarrow \mathcal{Z} \subseteq \mathbb{R}^{d_z}$ where $d_z$ is the length of each feature vector, and the discrimination function of the label classifier is $h: \mathcal{Z} \rightarrow  \mathcal{Y}$.

% In this subsection, we propose a more general domain adaptation problem GUDA. 
Given $\mathcal{Y}_c = \{1,2,\dots, k\}$ as the common label space between the source and target domain. We denote $\overline{\mathcal{Y}}_t = \{ k+1, \dots , k+n \}$ as the target private label space and $\overline{\mathcal{Y}}_s $ the source private label space, i.e., $\mathcal{Y}_s = \mathcal{Y}_c \cup \overline{\mathcal{Y}}_{s}, ~ \mathcal{Y}_t = \mathcal{Y}_c \cup \overline{\mathcal{Y}}_{t}$. Then we have GUDA in Definition~\ref{def-GUDA}.
\begin{definition}[GUDA] \label{def-GUDA}
    GUDA is characterized by conditions \ding{172} $\wedge$ \ding{173} $\wedge$ \ding{174} $\wedge$ \ding{175} $\wedge$ \ding{177}, i.e.,
\begin{equation}
\begin{aligned}
    P_s(X|Y)& \neq P_t(X|Y)>0, ~ P_s(Y) \neq P_t(Y) > 0 ,~ \exists Y \in \mathcal{Y}_c,  \\
   \text{and}&  \quad \overline{\mathcal{Y}}_s  \neq \emptyset,  \quad \overline{\mathcal{Y}}_t  \neq \emptyset,  \quad |\widehat{\mathcal{Y}}| = k+n,
   \end{aligned}
\end{equation}
where $\overline{\mathcal{Y}}_s$ and $\overline{\mathcal{Y}}_t$ are both unknown.
\end{definition}
% Overall, GUDA is a more comprehensive and realistic problem that encompasses variants in terms of label distribution, label space and label prediction space. 
% It is focused on training a model to predict target labels of all samples, including the private label set, while also accounting for unknown label spaces in both source and target domains and generalized label shift. 
GUDA focuses on predicting all target labels, including the private label set, while also accounting for unknown label spaces and generalized label shift. 
% GUDA aims to train a model $h$ to align correct labels for all target samples, including the target private label set.
 Therefore, the target risk of GUDA can be divided into two parts: the target common risk for classifying common classes and the refined target private risk for classifying the target private classes:
\begin{align}
    \label{eq:risk3}
    &\epsilon_t({h} \circ g) = \mathbb{E}_{({X}, {Y}) \sim \mathcal{D}_t} \ell({h} \circ g({X}), {Y})  \\
    &=\underbrace{\sum_{i=1}^{k} P_t(Y=i) \epsilon_{s, i}(h  \circ g)}_{\text{target common risk}} + \underbrace{\sum_{j=k+1}^{k+n}P_t(Y=j) \epsilon_{s, j}(h \circ g)}_{\text{refined target private risk}}. \nonumber
\end{align}
To minimize the target risk, an AL strategy can be employed to annotate a small portion of the target dataset to recover the target distribution. We denote $\mathcal{D}_{l}={(x_i, y_i)}_{i=1}^{b}$ as the selected labeled target dataset and the probability distribution of $\mathcal{D}_{l}$ as $P_{l}$.

\section{Method}
\label{sec:GFlowDA}

\subsection{Towards GUDA from a Theoretical View}
\label{theoretical}

 To start with, we provide a theoretical analysis of the proposed GUDA. First,
%   we introduce two related definitions:
 we introduce the definitions of two performance metrics for the predictor $h \circ g$:
\begin{definition}[Balanced Error Rate \cite{tachet2020domain}]
    Given a distribution $P_s$, the balanced error rate (BER) of a predictor $h \circ g$ on $P_s$ is given by:
$$
\varepsilon_{s}(\widehat{Y}||Y):=\max_{j \in \mathcal{Y}}P_s(\widehat{Y} \neq Y \mid Y=j),
$$
where $\widehat{Y} = h \circ g(X)$.
\end{definition}
% Specifically, for each class $j \in \mathcal{Y}$, BER represents the maximum probability that the predicted label $\widehat{Y} = h \circ g(X)$ is not equal to the true label Y, under the given distribution $P_s$.

\begin{definition}[Conditional Error Gap \cite{tachet2020domain}]
Given two distributions $P_s$ and $P_t$, the conditional error gap (CEG) of a classifier $h \circ g$ on $P_s$ and $P_t$ is given by
$$
\Delta_{s,t}(\widehat{Y}||Y):=\max_{j\neq i \in \mathcal{Y}}|{P}_s(\widehat{Y}=i|Y=j)-{P}_t(\widehat{Y}=i | Y=j)|.
$$
\end{definition}

BER measures max prediction error on $P_s$, reflecting the classification performance of a single domain. CEG characterizes the max discrepancy between the classifier's predictions on $P_s$ and $P_t$, reflecting the degree of conditional feature alignment across domains.

Then we present the target risk upper bound for GUDA in Theorem~\ref{theorem1} based on these two definitions, proved in the appendix.

\begin{theorem}[Target Risk Upper Bound for GUDA]\label{theorem1}
    Let $Y_c$, $Y_l$ and $\overline{Y}_t$ be random variables taking values from $\mathcal{Y}_c$, $\mathcal{Y}_l$ and $\overline{\mathcal{Y}}_t$ respectively, with $\mathcal{Y}_l$ being the selected target label space.
    For any classifier $\widehat{Y} = (h \circ g)(X)$, we have
\begin{align}
& \epsilon_t(h \circ g) 
\leq  \epsilon_{s}(h \circ g)+  \epsilon_{l}(h \circ g) + \delta_{s,t} + \delta_{l,t} +\varepsilon_{{t}}(\widehat{Y}||\overline{Y}_t), \nonumber
\end{align}
where 
\begin{align}
    \delta_{s,t} &= \left\|P_{s}(Y_c)-P_{t}(Y_c)\right\|_1 \cdot \varepsilon_{{s}}(\widehat{Y}||Y_c) + 2(k-1) \Delta_{{s}, t}(\widehat{Y}||Y_c),  \nonumber \\
    \delta_{l,t} &= \left\|P_{l}(Y_l)-P_{t}(Y_l)\right\|_1 \cdot \varepsilon_{{l}}(\widehat{Y}||{Y}_{l}) + 2(v-1) \Delta_{{l}, t}(\widehat{Y}||Y_l) \nonumber
\end{align}
with $k = |\mathcal{Y}_c|$ and $v = |\mathcal{Y}_l|$, $\|{P}_s(Y_c)-{P}_t(Y_c)\|_1= \sum_{i \in \mathcal{Y}_c}|P_s(Y=i)$ $-P_t(Y=i)|$ being the $L_1$ distance between $P_s(Y)$ and $P_t(Y)$ on the common label space $\mathcal{Y}_c$.
\end{theorem}

% \noindent
% \textbf{Remark}
\begin{remark}
The upper bound $\epsilon_t(h \circ g)$ contains five terms. 
The first two terms represent the source risk on $\mathcal{Y}_c$ and the selected target risk on $\mathcal{Y}_l$.
The third and fourth terms contain $\|{P}_s(Y_c)-{P}_t(Y_c)\|_1$ and $\|{P}_l(Y_l)-{P}_t(Y_l)\|_1$ respectively, which both measure the distances of the marginal label distributions across domains.  
The former is a constant that only depends on $\mathcal{D}_s$ and $\mathcal{D}_t$ while the latter changes dynamically since the construction of $\mathcal{D}_l$ varies with the AL strategy. 
$\varepsilon_a(\cdot)$ with $a \in \{s,l,t\}$ in the last three terms measure classification performance on the corresponding domains. 
$\Delta_{a,t}(\cdot)$ with $a \in \{s,l\}$ in the third and fourth terms measures the performance discrepancy between $P_a(Y)$ and $P_t(Y)$.
\end{remark}
The difficulty in minimizing the upper bound stems from the unknown nature of the target label distribution $P_t(Y_t)$ and the label space $\mathcal{Y}_t$.
One potential solution is to minimize the label distribution distance $|P_{l}(Y_l)-P_{t}(Y_l)|_1$ while making $|\mathcal{Y}_l|$ close to $|\mathcal{Y}_t|$. To achieve this goal, it is crucial to develop an effective AL strategy to select informative and representative samples for constructing $\mathcal{D}_l$ that preserves the entire target label distribution and label space. By doing so, $\mathcal{Y}_l$ can be considered a proxy of $\mathcal{Y}_t$ and $P_l$ a proxy of $P_t$. 
We propose GFlowDA based on a GFlowNet generator to generate $\mathcal{D}_l$ by sequentially adding one sample at each step until the budget is used up. 
We give the formal definition of GFlowDA in Definition ~\ref{definiton_GFlowDA}.

\begin{figure*}[tbp]
\begin{center}
\includegraphics[width=6.8in,height=2.6in]{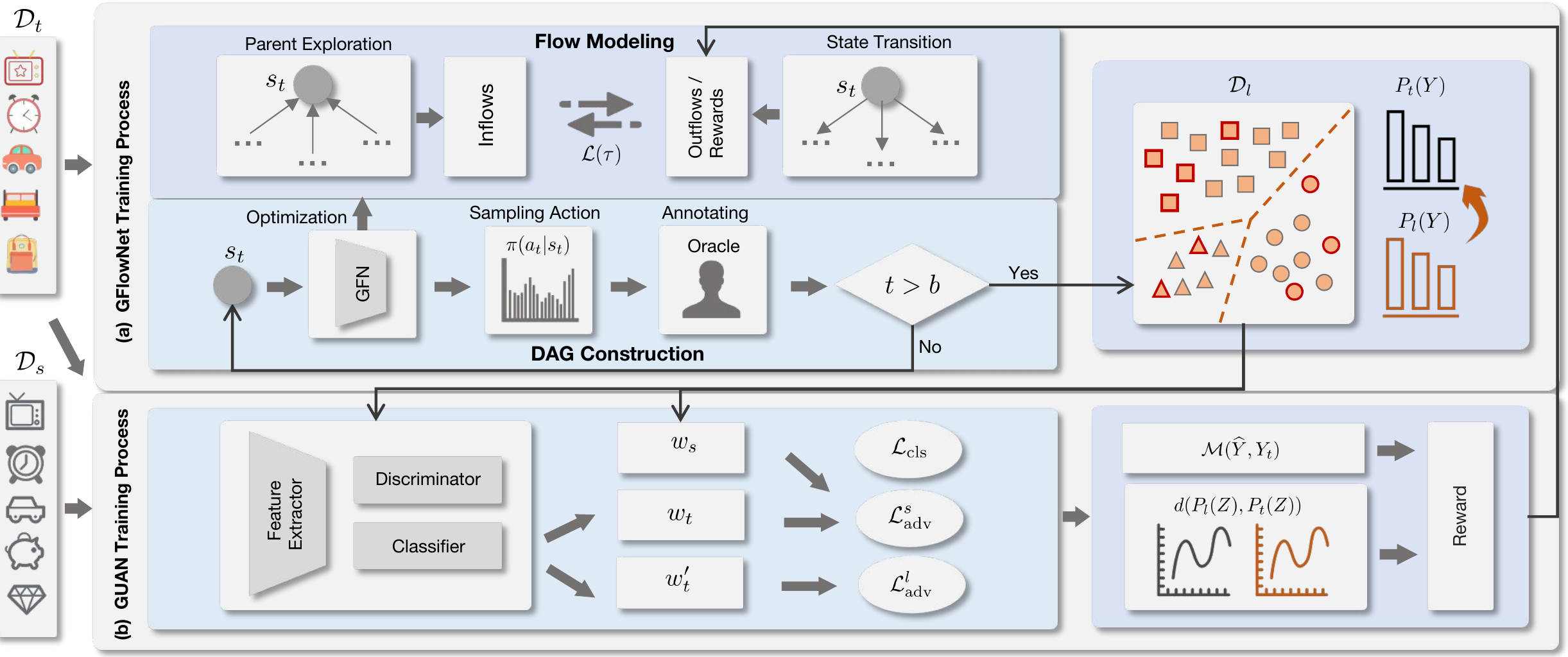}
\end{center}
\vspace{-0.2cm}
\caption{Overall framework of GFlowDA. GFlowNet selects and labels target samples, which are then fed as $\mathcal{D}_l$ to GUAN. GUAN is optimized based on $\mathcal{D}_l$, $\mathcal{D}_s$ and $\mathcal{D}_t$. The resulting reward is fed back to GFlowNet, which is then optimized by combining the reward with the inflows and outflows obtained through the parent exploration and state transition process.}
\label{fig:framework}
\vspace{-0.3cm}
\end{figure*}

 \begin{definition}[GFlowDA]
\label{definiton_GFlowDA}
Given a source distribution ${P}_s$ and a target distribution ${P_t}$, GFlowDA aims to find the best forward generative policy $\pi(\theta)$ based on flow network to generate a distribution $P_l$ automatically, which can serve as a proxy of $P_t$.  The probability of sampling the distribution $P_{l}$ satisfies
\begin{equation}
\label{eq_GFlowDA}
    \pi(P_{l}; \theta) \propto r(P_{s}, P_{t}, P_{l}),
\end{equation}
where $\theta$ is the parameter of flow network, $r(\cdot)$ is the reward function based on $P_s$, $P_t$ and $P_l$. 
% The learned policy $\pi$ can be transferred to the different target distributions.
\end{definition}

\subsection{DAG Construction of GFlowDA}
\label{sec:framework}

Consider a direct acyclic graph (DAG) $\mathcal{G}=(\mathcal{S}, \mathcal{A})$, where $\mathcal{S}$ and $\mathcal{A}$ are state/node and action/edge sets, respectively. Elements of them at step $t$ are denoted as $s_t$ and $a_t$ \footnote{Unless otherwise specified, the subscript "t" denotes "step" when used in conjunction with "s" and "a", and "target" when used in conjunction with "$x$" and "$\mathcal{D}$".}
The complete trajectory is a sequence of states $\tau=(s_0, \ldots, s_f)$.  
To construct $\mathcal{G}$,  we first define the state, action and reward function.

\begin{definition}[State]
    A state $s_t \in \mathcal{S} \subseteq \mathbb{R}^{n\times 4}$ in GFlowDA at step $t$ describes the entire target information based on $\mathcal{D}_t$ and the currently labeled data $\mathcal{D}_{l}$, where $s_t = \{s_t^i\}_{i=1}^{n}$ and $s_t^i$ is the state representation of the $i$-th target sample.
\end{definition}

We use $s^i_t(a), a \in \{0,1,2,3\}$ denotes the $a$-th column of the state matrix.
The first column of the state
denotes the maximum similarity between target features and selected target features at step $t$. Intuitively, selecting samples with low maximum similarity values can ensure the  instance-level diversity, i.e.,
\begin{equation}
\label{state1}
\begin{aligned}
s^i_t(1) = 
 \operatorname{max}_{x_j \in \mathcal{D}_l} \operatorname{cos}(g(x_i),g(x_j)).
\end{aligned}
\end{equation}
The second column of the state denotes the maximum similarity between target features and active target prototypes, which ensures class-level diversity, i.e.,
\begin{equation}
\label{state2}
\begin{aligned}
s^i_t(2) = 
\operatorname{max}_{j \in \mathcal{Y}_l} \operatorname{cos}(g(x_i),{\mu_{l}^j}),
\end{aligned}
\end{equation}
where  $\mu_{l}^j$ is the prototype of class $j$ in $\mathcal{D}_l$, calculated as follows: 
\begin{equation}
 \mu_{l}^{j}=\frac{\sum_{(x,y) \in \mathcal{D}_{l}} \mathds{1}\{y = j\}  g({x})}
 {\sum_{(x,y) \in \mathcal{D}_{l}} \mathds{1}\{y = j\}}  .
 \label{muk}
\end{equation}
The third column of the state denotes the uncertainty of the target samples, which is calculated by the entropy of the label probabilities $\hat{y}_i = h \circ g(x_i)$, i.e.,
\begin{equation}
\label{state3}
\begin{aligned}
s^i_t(3) = 
H(\hat{y}_i)).
\end{aligned}
\end{equation}
The last column of the state is an indicator variable to represent whether a sample has been labeled or not, i.e., 
\begin{equation}
\label{state4}
\begin{aligned}
s^i_t(4) = \mathds{1}\{x_i \in \mathcal{D}_{l}\}.
\end{aligned}
\end{equation}

% the similarity relation between active prototypes $\mathbf{\mu}_t$ and target instances $\mathcal{D}_T$ and whether a target instance has been labeled or not. 

\begin{definition}[Action]
    An action $a_t \in \mathcal{A} $ at step $t$ in GFlowDA determines which target instance will be selected from candidate target dataset $\mathcal{D}_t \backslash \mathcal{D}_{l}$. 
\end{definition}
% \textbf{Action:} An action $a_t \in \mathcal{A} $ at step $t$ in GFlowDA determines which target instance will be selected from candidate target dataset $\mathcal{D}_t \backslash \mathcal{D}_{l}^t$ with $\mathcal{D}_{l}^t$ being the selected labeled samples up to step $t$. 

% At each step $t$, the agent decides the action to take based on its policy $\pi(a_t|s_t)$. Therefore the $a_t$-th instance of the unlabelled target dataset will be selected and queried by human oracle. Once the sample is selected, the agent is unable to choose it again in the subsequent steps.

\begin{definition} [Reward Function]
\label{def:reward}
    A reward function $r(s_f)$ of the terminal state $s_f$ in GFlowDA refers to a comprehensive metric measuring the quality of $\mathcal{D}_{l}$, by taking into account the diversity and informativeness, which is expressed as:
    \begin{equation}
        \label{reward}
        \begin{aligned}
            r(s_f) =& 
             - \operatorname{MMD}(g(X_t), g(X_l) + \mathcal{M}(\widehat{Y}, Y_t) ,
        \end{aligned}
        \end{equation}
        where $\widehat{Y} = h \circ g (X)$, $\operatorname{MMD}(\cdot)$ represents the Maximum Mean Discrepancy (MMD)~\cite{long2017deep} between the source and target marginal feature,  and $\mathcal{M}(\cdot)$ means the classification accuracy used to evaluate the performance of $h \circ g$.
\end{definition}

% \noindent
% \textbf{Remark}
\begin{remark}
MMD distance can encourage diverse sample selection that preserves the entire target distribution. To make the selected samples more informative, we incorporate the classification accuracy of the model on the target domain as part of the reward. 
\end{remark}

\subsection{Flow Modeling of GFlowDA}
\label{sec:efficent}

\begin{figure}[tbp]
    \begin{center}
    \includegraphics[width=3.3in]{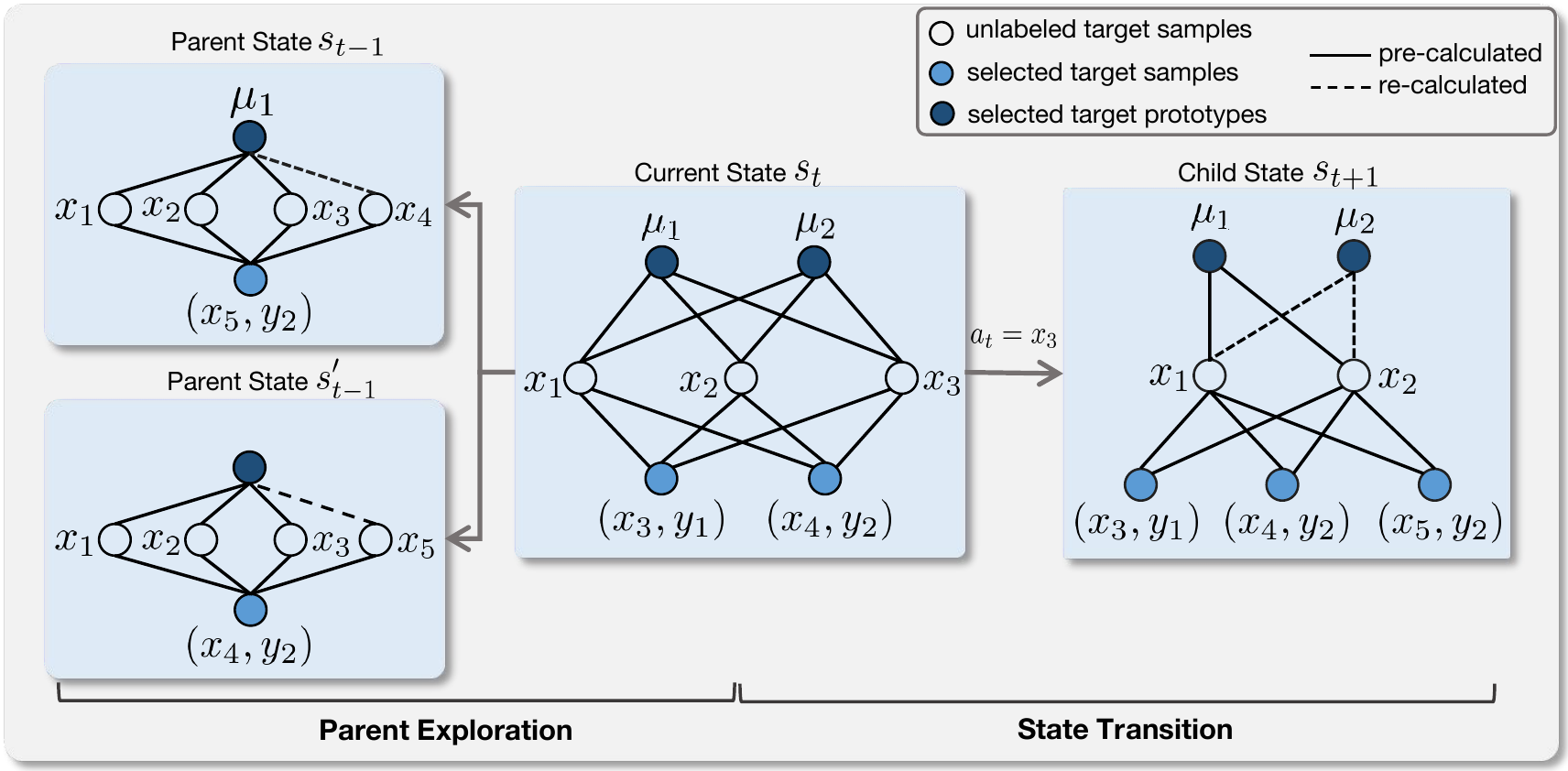}
    \end{center}
    \vspace{-0.2cm}
    \caption{Illustration of efficient parent exploration and state transition. 
    $\textbf{Left:}$ At state $s_t$, $x_3$ and $x_4$ are labeled as $y_1$ and $y_2$, resulting in two parent states. 
    \textbf{Right:} Due to $a_t = x_3$ , $x_3$ is labeled at $s_{t+1}$, and the prototype $\mu_2$ needs to be updated. 
    }
    \label{fig:state_transition}
    \vspace{-0.3cm}

\end{figure}

To achieve Eq.~\ref{eq_GFlowDA}, a non-negative function $F(\cdot)$ is introduced to measure the probabilities associated with $s_t$, where $F(s_t, a_t)=F(s_t \rightarrow s_{t+1})$ corresponds to an action/edge flow. The trajectory flow is denoted as $F(\tau)$ and the state flow is the sum of all trajectory flows passing through that state, denoted as $F(s)=\sum_{s \in \tau} F(\tau)$. 
Our objective is to ensure that the DAG $\mathcal{G}$ operates analogously as a water pipe, where water enters at $s_0$ and flows out through all $s_f$, satisfying the condition $F(s_0)=\sum F(s_f)=\sum r(s_f)$.  To achieve this, we need to calculate the inflows and outflows of each state, corresponding to the parent exploration and state transition, respectively.
 As illustrated in Fig.~\ref{fig:state_transition}, in the parent exploration process, we explore all direct parent states of $s_t$, i.e., $s \in \mathcal{S}_p(s_{t})$ with $\mathcal{S}_p(s_{t})$ being the parent set. We have the following proposition to guide the exploration procedure:
\begin{proposition}
\label{proposition}
    For a state $s_{t}$ in GFlowDA, the number of its parent nodes is equal to the number of labeled samples currently, i.e.,
$|\mathcal{S}_p(s_{t})| = ||s_{t}(4)||_0$.
\end{proposition}
% For a state $s_{t}$ in GFlowDA, the number of its parent nodes is equal to the number of labeled samples currently, i.e.,
% $|\mathcal{S}_p(s_{t})| = ||s_{t}(4)||_0$.
Based on Proposition~\ref{proposition}, one of $s_{t}$'s parents is obtained by three steps: (a) selecting the $j$-th element in the vector $s_t(4)$ and setting its value from $1$ to $0$ to obtain a new vector $s_{t-1}(4)$, (b) updating the instance-level similarity and (c) updating the class-level similarity based on $s_{t-1}(4)$. 
Notably, $s_t(3)$ remains fixed as the domain adaptation model updates only at the terminal state. 
However, directly computing all parent states using Eq.\ref{state1} and Eq.\ref{state2} is too time-consuming. Therefore, we propose a more efficient approach to update the similarities.  
For instance-level similarity, we pre-calculate a cosine similarity matrix $\text{SIM}(X_t)$ between all target domain samples at the start of the generative process. The state of the $i$-th target sample in the first column can be updated as follows.
\begin{equation}
\label{eq:s1_update}
    s_{t-1}^i(1) = \max(s^i_t(1), \text{SIM}_{i,j}),
\end{equation}
where $\text{SIM}_{i,j}$ denotes the similarity between the $i$-th target sample and the $j$-th target sample.
For class-level similarity, we just need to update $\mu_t^{c}$ with $c$ being the label of the selected $j$-th sample. $\mu_{t-1}^{c}$ can be quickly calculated based on $\mu_{t}^{c}$, which is given by
\begin{equation}
\label{eq:miu_update1}
    \mu_{t-1}^{c} = \frac
    {\mu_{t}^{c} \cdot (\sum_{{x} \in \mathcal{D}_{l}} \mathds{1}\{y =c\}) - g(x_j)}
    {\sum_{{x} \in \mathcal{D}_{l}} \mathds{1}\{y = c\} - 1}.
\end{equation} 
Then the state of the $i$-th target sample in the second column can be updated as follows:
\begin{equation}
    \label{eq:s2_update}
    s_{t-1}^i(2) = \max(s^i_t(2), \cos( \mu_{t-1}^{c},x_i)).
\end{equation}
% where $\cos(\cdot)$ is the cosine similarity operator.

In the state transition process, if $s_t$ is not a terminal state, we update the dynamic features of $s_t$ to obtain its child state $s_{t+1}$. 
This process can be considered an inverse repetition of the parent exploration process.
Specifically, once $a_t = j$ is sampled, the $j$-th element in vector $s_t(4)$ is updated from $0$ to $1$, resulting in a new vector $s_{t+1}(4)$. Based on $s_{t+1}(4)$,$s_{t+1}(1)$ can be updated by utilizing Eq.~\ref{eq:s1_update}. 
To update $s_{t+1}(2)$, we use a similar equation as Eq.\ref{eq:s2_update}. Since we need to add a new sample instead of removing one as in Eq.\ref{eq:miu_update1}, $\mu_{t+1}^{c}$ is updated as follows:
\begin{equation}
    \label{eq:miu_update2}
        \mu_{t+1}^{c} = \frac
        {\mu_{t+1}^{c} \cdot (\sum_{{x} \in \mathcal{D}_{l}} \mathds{1}\{y = c\}) + g(x_j)}
        {\sum_{{x} \in \mathcal{D}_{l}} \mathds{1}\{y = k\} + 1}.
    \end{equation}

Overall, efficient exploration reduces complexity and accelerates the training of the generative policy network with flow matching loss (discussed in the next subsection).

\subsection{Training Procedure}
\label{sec:training}
As illustrated in Fig.~\ref{fig:framework}, GFlowDA consists of two primary components: a generative policy network for selecting and annotating the most useful target samples, and a domain adaptation network that utilizes the annotated samples to adapt the target domain.

\noindent
\textbf{Genrative Policy Network.} 
Based on the parent set $\mathcal{S}_p(s_{t})$ and child set $\mathcal{S}_c (s_{t})$ obtained through the above exploration and transition process, we can calculate the corresponding \textbf{\emph{inflows}} and \textbf{\emph{outflows}}. The inflows are calculated by 
% $
% \sum_{s_{t-1},a_{t-1}:T(s_{t-1},a_{t-1}) = s_{t}} F(s_{t-1},a_{t-1})
% $,
$
\sum_{s \in \mathcal{S}_p(s_{t})} F(s)
$,
which represents the sum of action flows from all parent states. 
The outflows are flows passing through it, which is given by 
% $
% \sum_{a_{t} \in \mathcal{A}}F (s_{t},a_{t})
% $.
$
\sum_{s \in \mathcal{S}_c (s_{t})} F(s)
$.
Starting from an empty set, GFlowDA draws complete trajectories $\tau=(s_0, s_1, \ldots, s_f)$ by iteratively sampling target samples until the budget is used up. After sampling
a buffer, we train the policy $\pi(\theta)$ to satisfy Eq.~\ref{eq_GFlowDA}, by minimizing the loss over the flow matching condition:
\begin{equation}
    \begin{aligned}
    \label{gflow_loss}
    & \mathcal{L}(\tau)  =\sum_{s_{t} \in \tau \neq s_0}\Bigg\{ \log \bigg[\epsilon+\sum_{s \in \mathcal{S}_p(s_{t})} F^{\prime} \bigg]  \\
    & - \log   \bigg[\epsilon + \mathds{1}_{s_{t}=s_f}r\left(s_{f}\right)+  \mathds{1}_{s_{t} \neq s_f} \sum_{s \in \mathcal{S}_c(s_{t})} F^{\prime} \bigg] \Bigg\}^2,  
    \end{aligned}
    \end{equation}

where $F^{\prime} = \exp(\log F(s)$), and the reward $r(s_f)$ is computed by Eq.~\ref{reward}. 
For internal states, we only calculate outflows based on action distributions. For terminal states, we calculate their rewards to evaluate the efficacy of $\mathcal{D}_l$.
To acquire a dependable reward, we introduce a new domain adaptation network as follows.

\begin{table*}[t]
    \caption{Average class accuracies (\%) and JSD (\%) on \textbf{Office-31} with 5\% target data as the labeling budget.}
    \vspace{-0.2cm}
    \label{tab:table_office}
    \centering
    \tiny
    \resizebox{1\linewidth}{!}{
    \begin{tabular}{l|cccccccccccc|cc}
    
    \toprule
    \multirow{2}{*}{Method} &    \multicolumn{2}{c}{A $\rightarrow$ D} &
    \multicolumn{2}{c}{A $\rightarrow$  W} &
    \multicolumn{2}{c}{D $\rightarrow$  A} &
    \multicolumn{2}{c}{D $\rightarrow$  W} &
    \multicolumn{2}{c}{W $\rightarrow$  A} &
    \multicolumn{2}{c|}{W $\rightarrow$  D} &
    \multicolumn{2}{c}{Avg} \\  
    
        &Acc        &JSD        &Acc        &JSD         &Acc        &JSD         &Acc        &JSD        &Acc        &JSD        &Acc        &JSD      &Acc        &JSD        \\ 
    % Methods & Art & Cartoon & Photo & Sketch & Average \\
    \midrule
    {Random}      &   46.23   &   16.22  &   60.79   &  {15.80}  & 55.67     &3.14  &  67.98  &  \underline{14.39}    &   61.80  &    3.57  & 67.46   &   \underline{15.64}   &   59.99 &  \underline{11.46}                 \\
{Entropy~\cite{wang2014new}}          &   53.85   & 27.93    &   57.12   &  23.53  &   65.88   & 15.09  & 70.12   &  22.96    &   \underline{67.16}  & 4.93     & 70.91   & 35.04     &   64.17   & 21.58              \\
{TQS~\cite{fu2021transferable}}  &      63.10       &   \underline{16.02}  &    67.80     &  15.92&   65.72 & \underline{2.67}   &  \underline{82.40}    &   16.84   &   66.85    &\underline{3.01} &  \underline{80.65}    &   21.93  &   71.09    & 12.73                           \\
{CLUE~\cite{prabhu2021active}}    &56.89     &   23.96   &  66.12 & \underline{15.36}     &   61.01 &3.97  &81.37      & 16.39 &   60.13&  7.23   & 75.27   &  22.06     &   66.80   & 14.83                                 \\
{EADA~\cite{xie2022active}} &   46.63   &  27.15   & 65.40    &    21.10    &    51.05     &  7.60  &     72.40    &    28.39    & 54.22     &7.11 &  73.02    &  22.93    &  60.45  & 19.05                    \\
{SDM-AG~\cite{xie2022learning}} &  62.76  &   22.85  &   68.20 &  17.38     &  65.3& 5.67   &   79.40 &   21.42   &     64.41 &  3.49&  72.43    &   25.87  &  68.75  & 16.11                                          \\    
{AUDA~\cite{xie2022active}}   &   65.13   &  21.36   &    67.36  & 18.32   &  67.12    & 5.01 & 79.60   & 20.87     &  63.78   &  4.12    & 77.14   & 19.36     &  70.02    & 14.84                \\         
% {LAMDA~\cite{hwang2022combating}}   &   65.13   &  21.36   &    67.36  & 18.32   &  67.12    & 5.01 & 79.60   & 20.87     &  63.78   &  4.12    & 77.14   & 19.36     &  70.02    & 14.84                \\   
\midrule
{RLADA (Ours)} &  \underline{67.36} &     20.33 &   \underline{71.23}  &   16.79   & \underline{70.56}   &  3.67    &  79.88  & 16.03   &   65.98   &   3.20  &   76.78   &  19.66  &  \underline{71.96}   &  13.28 \\ 
{GFlowDA (Ours)} &   \textbf{70.50}  &   \textbf{15.01}   &   \textbf{73.00}    & \textbf{13.54}    & \textbf{74.80}   & \textbf{1.72}   &  \textbf{82.60}    &   \textbf{11.93}   &  \textbf{67.51}    & \textbf{1.51} &    \textbf{80.75}   &    \textbf{14.40} &    \textbf{74.86}  &  \textbf{9.68}         \\
    \bottomrule
    \end{tabular}}
    \vspace{-0.2cm}
    \end{table*}

\begin{table*}[t]
    \caption{Average class accuracies (\%) and JSD (\%) on \textbf{Office-Home} with 5\% target data as the labeling budget. }
    \vspace{-0.2cm}
    \label{tab:table_oh}
    \centering
    \tiny
    \resizebox{1\linewidth}{!}{
    \begin{tabular}{l|cccccccccccc|cc}
    
    \toprule
    \multirow{2}{*}{Method} &    
    \multicolumn{2}{c}{Ar$\rightarrow$ Cl} &
    \multicolumn{2}{c}{Ar$\rightarrow$ Pr} &
    \multicolumn{2}{c}{Ar$\rightarrow$ Rw} &
    \multicolumn{2}{c}{Cl$\rightarrow$ Ar} &
    \multicolumn{2}{c}{Cl$\rightarrow$ Pr} &
    \multicolumn{2}{c|}{Cl$\rightarrow$ Rw} &
    \multicolumn{2}{c}{Avg (12 tasks)}  \\  
    
        &Acc        &JSD        &Acc        &JSD         &Acc        &JSD         &Acc        &JSD        &Acc        &JSD        &Acc        &JSD      &Acc        &JSD        \\ 
    % Methods & Art & Cartoon & Photo & Sketch & Average \\
    \midrule
    {Random}        &   38.80  &  7.24   &   63.92   & 6.06   &    63.67  & 6.91 &    42.38&  12.14    &  61.69   &  5.47    & 65.48   &   \underline{5.40}  &  52.20 &  7.76                 \\
{Entropy~\cite{wang2014new}}           &   37.39   &  8.79   &  65.79    & 7.20   &  57.75    & 22.09 &  38.52  &  14.70    &   56.22  &  10.92    &  55.37  &    12.49   & 48.36   &   14.72            \\
{TQS~\cite{fu2021transferable}} &  43.55    & 6.15    &  68.94    &  8.38   & 64.47 & 5.37 &  39.39  & 11.37     &59.63   &   \textbf{6.12}   &  54.00  & 7.74     &   52.49  & 8.04                             \\
{CLUE~\cite{prabhu2021active}}  &   42.14   &   7.12  &  70.12    & \underline{5.88}   &   65.12   & \textbf{4.78}  &  41.57  & \underline{7.54}     & 63.11    &   7.23 & 56.23  &  7.17    &  53.98 &  \underline{7.16}           \\
{EADA~\cite{xie2022active}} &  40.36    &  \underline{5.65}   &    69.06  & 7.44   & 62.19     & 6.17 &  30.67  &  10.68    &  65.14   &   9.47   &   54.53 &   7.64 & 50.28 & 9.32         \\
{SDM-AG~\cite{xie2022learning}} &    40.01  & 8.43    &  69.25    & 7.99   &   62.26   & 8.54 & 38.21  & 9.74     &    53.13 & 10.98     &  51.51  & 9.78 & 42.84   &10.83             \\      
{LAMDA~\cite{hwang2022combating}}   &   42.19   &  8.59   &  59.75    &   7.22 &       
\underline{70.51}    &9.25  & \underline{44.80}   & 17.49     &  \underline{67.17}   &    6.21  & \underline{52.47} &   10.21   &  
53.86  &  11.37             
\\   
\midrule
{RLADA (Ours)} & \underline{47.56}   &   7.12   &    \underline{72.69} & 6.73  &  65.81  &  7.33    &40.37  & 10.67   &     66.52 &  8.64   &  60.46    & 6.48   & \underline{55.32}   &   7.68\\ 
{GFlowDA (Ours)} & \textbf{51.89}   &   \textbf{5.01}  &     \textbf{74.77} &  \textbf{4.87}  &  \textbf{72.99}  & \underline{5.46}  &   \textbf{44.23} &  \textbf{7.12}    &   \textbf{72.29}  &  \underline{6.18}    &  \textbf{67.11}  & \textbf{5.03}&   \textbf{59.32} &  \textbf{5.65}       \\
    \bottomrule
    \end{tabular}}
    \vspace{-0.2cm}
    \end{table*}

\noindent
\textbf{Generalized Universal Adversarial Network.} 
In the GFlowDA framework, we propose a novel weighted adaptive model for GUDA named Generalized Universal Adversarial Network (GUAN).
In addition to the feature extractor $g$ and classifier $h$, GUAN also includes a domain discriminator $d: \mathcal{Z} \rightarrow \mathbb{R}$, which aims to align the source and target features adversarially.
The weighted alignment losses of source data and labeled target data are formally described as follows
\begin{equation}
\label{da_loss1}
\mathcal{L}^{s}_{\text{adv}}=-\mathbb{E}_{x \sim P_{s}} {w}_{s}  \log \left(d(z)\right) - \mathbb{E}_{x \sim P_t} {w}_t   \log \left(1- d(z)\right)
\end{equation}
\begin{equation}
\label{da_loss2}
\mathcal{L}^{l}_{\text{adv}}=-\mathbb{E}_{x \sim P_{l}}   \log \left(d(z)\right)- \mathbb{E}_{x \sim P_{t}} {w}'_{t}   \log \left(1- d(z)\right) ,
\end{equation}
% where $\hat{z} = d \circ g\left(x\right)$, $d$ is a discriminator, $\mathbf{w}_s$, $\mathbf{w}_t$ and $\mathbf{w}'_t$ are defined as 
where $z = g(x)$. $w_s$ in Eq.~\ref{da_loss1} indicates the ratio between the estimated target label distribution and the source label distribution if $x$ is determined to belong to  $\mathcal{Y}_c$, which is defined as follows:
\begin{align}
    & {w}_s = \mathds{1}(y \in \mathcal{Y}')\cdot (P_l(y) / P_s(y)) + (1 - \mathds{1}(y \in \mathcal{Y}')) \cdot \lambda \nonumber
 \end{align}
 where $\mathcal{Y}' = \mathcal{Y}_l \cap \mathcal{Y}_s$ acting as an approximation of the common label space $\mathcal{Y}_c$, and $\lambda$ is a constant working as a remedy for compatibility with the potential inconsistency between $\mathcal{Y}'$ and $\mathcal{Y}_c$.

$w_t({x})$ in Eq.~\ref{da_loss1} and $w'_t({x})$ in Eq.~\ref{da_loss2} indicate the probability of a target sample ${x}$ belonging to the source label set $\mathcal{Y}_s$ and the selected labeled target label set $\mathcal{Y}_l$, respectively. 
 We use the predicted probabilities over all labels in $\mathcal{Y}_s$ and $\mathcal{Y}_l$ respectively to estimate the probabilities, as described below:
\begin{equation}
    {w}_t =\frac{1}{u} \sum_{i \in \mathcal{Y}_s} \hat{\bm{y}}[i] ,\quad {w}'_t =\frac{1}{v} \sum_{i \in \mathcal{Y}_l} \hat{\bm{y}}[i]
\end{equation}
where $k = |\mathcal{Y}_c|$ and $v = |\mathcal{Y}_l|$, $\hat{\bm{y}} = h(z)$ represents the prediction probability vector.

% ${w}_s$, ${w}_t$, and ${w}'_t$ are weighting coefficients, defined in equations (3)-(5). These weighting coefficients allow us to account for the presence of unknown and known target categories in the labeled and unlabeled target domains, respectively.

% \begin{equation}
%     \label{eq:w_t}
%     {w}_t = \frac{\sum^{m}_{i=1} \mathds{1}(i \in {\mathcal{Y}s}) \cdot \hat{\bm{y}}[i]}{\sum^{m}_{i=1}\mathds{1}(i \in {\mathcal{Y}_s})}
% \end{equation}

% where $\mathcal{Y}' = \mathcal{Y}_l \cap \mathcal{Y}_s$;  $\mathbf{w}_s$ is the label ratio if the source sample belongs to the estimated common label space $\mathcal{Y}'$, otherwise it is a constant $\lambda$; $\mathbf{w}_t$ and $\mathbf{w}'_t$ indicate the probability that the target sample belongs to $\mathcal{Y}_s$ or $\mathcal{Y}_l$, respectively.
% The classification loss is given by
The classification loss for GUAN aims to minimize the cross-entropy loss for both source and labeled target samples:
\begin{equation}
\label{da_loss3}
\begin{aligned}
\mathcal{L}_{\text{cls}}=&-\mathbb{E}_{x \sim P_{s}} L \left(\hat{y},{y}\right) 
-\mathbb{E}_{x \sim P_{l}} L\left(\hat{y},y\right) ,
\end{aligned}
\end{equation}
where $\hat{y} = h \circ g\left(x\right)$ and $L(\cdot)$ is the cross entropy loss.

% In GFlowDA, the generative policy network is updated through the reward computed by GUAN and then generates new selected target dataset $\mathcal{D}_l$, which in turn updates GUAN. This reciprocal relationship between the generative policy network and GUAN makes it possible to achieve efficient alignment between domains while improving the classification accuracy of the GUDA task.

\section{Experiments}
\label{sec:experiment}

\subsection{Experimental Setup}
% \label{dataset}
\textbf{Datasets.} 
We perform experiments on five benchmarks including Office-31 \cite{saenko2010adapting}, Office-Home \cite{venkateswara2017deep}, PACS \cite{li2017deeper} and VisDA~\cite{peng2018visda}.  To evaluate our algorithm on GUDA, we modify the source and target dataset by combining two subsampling protocols ~\cite{you2019universal,tachet2020domain}. These protocols are tailored for GLS and UniDA respectively. See the appendix for more details.

\begin{table*}[t]
    \caption{Average class accuracies (\%) and JSD (\%) on \textbf{PACS} and  with 1\% target data as the labeling budget.}
    \vspace{-0.2cm}
    \label{tab:table_pacs}
    \centering
    \tiny
    \resizebox{1\linewidth}{!}{
    \begin{tabular}{l|cccccccccccc|cc}

        % \multirow{2}{*}{Methods} & \multicolumn{5}{c||}{PACS dataset (\%)} & \multicolumn{5}{c}{Office-Home dataset (\%)} \\
        % \cline{2-11}
        % & Art & Cartoon & Photo & Sketch & Average & Art & Clipart & Product & Real-World & Average \\
    
    \toprule
    \multirow{3}{*}{Method} &  \multicolumn{12}{c|}{PACS (\%)} & \multicolumn{2}{c}{VisDA (\%)}\\
    % \cline{2-14}
    & 
    \multicolumn{2}{c}{A $\rightarrow$ C} &
    \multicolumn{2}{c}{A $\rightarrow$ P} &
    \multicolumn{2}{c}{A $\rightarrow$ S} &
    \multicolumn{2}{c}{C $\rightarrow$ A} &
    \multicolumn{2}{c}{C $\rightarrow$ P} &
    \multicolumn{2}{c|}{Avg (12 tasks)} &
    \multicolumn{2}{c}{S$\rightarrow$ R} \\  
    
        &Acc        &JSD        &Acc        &JSD         &Acc        &JSD         &Acc        &JSD        &Acc        &JSD        &Acc        &JSD      &Acc        &JSD        \\ 
    % Methods & Art & Cartoon & Photo & Sketch & Average \\
    \midrule
    {Random}       & 52.80     &  \textbf{0.29}   &   81.86   & 9.69   &   60.24   & 1.58 &  \underline{58.35}  &  \textbf{0.02}    & 82.80    & 11.98       & 63.27   &   4.30   &  55.52  &  \textbf{0.46}             \\
{Entropy~\cite{wang2014new}}      &    44.53  &  3.05   & 67.01     & 15.58   & 60.65     & \underline{1.42} &  35.71  & 5.99     &  84.44   &  16.94   & 58.38   & 8.20     & 50.45  & 8.42            \\
{TQS~\cite{fu2021transferable}}   &  63.82    &5.82     &   83.89   &  5.71  &    \underline{80.43}  & 2.77 &  44.48  & 4.39     &  87.65   &   3.47  &   63.12 & 6.37 &    82.87 &  5.10                      \\
% {CLUE~\cite{prabhu2021active}}      &   35.18   & 9.01    &    28.46  & 34.67   &   38.21   & 5.42 &  23.97  &   7.67   &  40.66   &  26.67    &33.10   &    5.80 &  27.09  & 7.17                \\
{EADA~\cite{xie2022active}}  &   35.18   & 9.01    &    28.46  & 34.67   &   38.21   & 5.42 &  23.97  &   7.67   &  40.66   &  26.67    &  34.38  &  11.16    &   \underline{84.73}   & 1.98               \\
{SDM-AG~\cite{xie2022learning}} &  \textbf{68.76}    &  3.91   &   \underline{90.63}   & 7.81   &    \textbf{84.03}  & 2.39 &   51.63 & 2.55     &   74.90  & 6.73     &63.16    & 4.18  & 80.69   & 2.60                   \\    
{LAMDA~\cite{hwang2022combating}}    &   53.20   & 1.62    &      82.67&  6.66  &    55.87  &2.96  &  \textbf{58.81}  & 1.04     &   \textbf{91.08}  &   9.81  &  65.97  & 5.85&  -   &   -         
\\   
\midrule
{RLADA (Ours)} & 59.78     &  3.28   &  90.12    & \underline{4.32}   &   67.03   &0.80  & 57.03   &    2.78  &  87.21   &    4.07  &  \underline{66.00}  & \underline{2.94}  & 81.23  & 3.23         \\ 
{GFlowDA (Ours)}   &    \underline{64.34}  &  \underline{1.06}   &    \textbf{91.38}  &  \textbf{4.11}  & 67.23     & \textbf{0.55} &   56.66 &   \underline{0.61}   &   \underline{90.54}  & \textbf{1.87}   &  \textbf{68.26}  & \textbf{1.93} &   \textbf{85.23}  & \underline{1.03}              \\
    \bottomrule
    \end{tabular}}
    \vspace{-0.3cm}
    \end{table*}
    
\begin{figure*}[htbp]
    \begin{center}
    \includegraphics[width=6.8in]{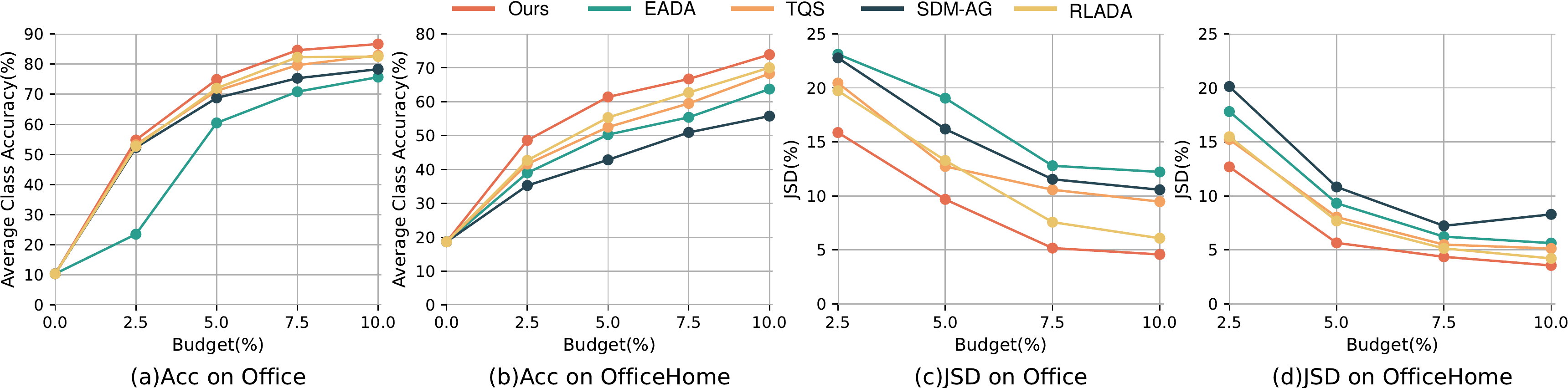}
    \end{center}
    \vspace{-0.2cm}
        \caption{Average class accuracies (\%) and JSD (\%) on Office-31 and Office-Home with different budget.}
    \label{fig:budget}
    \vspace{-0.2cm}
    \end{figure*}

% \textbf{Comparison Baselines.} 
% We compare GFlowDA against several AL and ADA methods including
% (1) Random, (2) Entropy~\cite{wang2014new}, (3) TQS~\cite{fu2021transferable}, (4) CLUE~\cite{prabhu2021active}, (5) EADA~\cite{xie2022active}, (6) SDM-AG~\cite{xie2022learning}, (7) LAMDA~\cite{hwang2022combating}. 
% Furthermore, to illustrate the superiority of GFlowDA, we also implemented a new algorithm by changing the policy network from GflowNet to Proximal Policy Optimization~\cite{schulman2017proximal} and keeping everything else the same, named (8) Reinforcement Learning Active Domain Adaptation, abbreviated as RLADA.

\textbf{Evaluation metrics.}
Besides reporting the average class accuracy, we also compared the Jensen-Shanon divergence (JSD) between the label distribution of the selected samples and the original target samples, denoted by $d_{\text{JS}}(P_l^Y, P_t^Y)$. 
JSD can provide an intuitive reflection of the label distribution reduction and exploration ability of AL strategies. A smaller JSD value indicates that the selected samples are better able to estimate the original distribution.
Compared baselines and Implementation details are introduced in Sec~\ref{comp} and Sec~\ref{implem} in the Appendix.

\subsection{Main Results of GFlowDA}

\textbf{Performance of GFlowDA on GUDA.}
The experimental results of different methods on Office-31, Office-Home, PACS and VisDA are shown in Table~\ref{tab:table_office}, Table~\ref{tab:table_oh} and Table~\ref{tab:table_pacs} respectively,
demonstrating that GFlowDA surpasses all the baselines by a large margin in both accuracy and JSD. 
It is worth noting that the random strategy outperforms most heuristic rule-based methods in terms of JSD, which is statistically intuitive that random selection is an independent and identically distributed (i.i.d.) procedure. However, random selection is unstable, which may explain why it achieves comparable JSD but 14.87\% lower accuracy than GFlowDA on Office-31.
Furthermore, two learning-based active strategies, GFlowDA and RLADA, achieve optimal and suboptimal accuracy on Office-31, Office-Home and PACS, indicating that such strategies have stronger exploration ability. 
However, it should be noted that GFlowDA can achieve better performance and exploration than RLADA. 
% This is because RLADA tends to explore near the maximum reward, while GFlowDA explores according to the reward distribution. 

\begin{table}[tbp]
    \caption{Comparison results on more settings and methods. }
    \vspace{-0.2cm}
    \begin{center}
    \resizebox{3.3in}{!}{
    \begin{tabular}{ccccccccc}
    \toprule
    \multirow{2}{*}{Method} 
     &
      \multicolumn{2}{c}{Office31-Origin} &
      \multicolumn{2}{c}{OH-RUST} &
      \multicolumn{2}{c}{Office31-GUDA} &
      \multicolumn{2}{c}{OH-GUDA}\\  
    \cmidrule(r){2-3}\cmidrule(r){4-5}\cmidrule(r){6-7} \cmidrule(r){8-9} 
    % \noalign{\smallskip}
       &Acc        &JSD         &Acc        &JSD         &Acc        &JSD  &Acc        &JSD       \\ 
    % \noalign{\smallskip}
    \midrule
    % \noalign{\smallskip}
    % SDM-AG  & \underline{92.00} & \textbf{14.40}  &  60.02  & 16.37  & 68.75 & 16.11 & 42.84  & \underline{10.83}    \\
    AUDA  &   90.43  & 16.32  & 59.43 & 14.85 & 66.53  & 17.35 & 50.35 & 12.10       \\
    LAMDA & 89.91  & 15.97  &  \underline{61.26}  & \underline{13.07}  & \underline{70.02}  & \underline{14.84} & \underline{53.86}  & 11.37    \\
    
    % \noalign{\smallskip}
    \midrule
    % \noalign{\smallskip}
    GFlowDA  & \textbf{92.85}  & \underline{15.67}  & \textbf{63.19} & \textbf{11.34}  &  \textbf{74.86} & \textbf{9.68} & \textbf{61.40} & \textbf{5.65}   \\
    % \noalign{\smallskip}
    \bottomrule
    \end{tabular}}
    \label{table:exist}
    \end{center}
    \vspace{-0.2cm}
    \end{table}

\textbf{Transferability of GFlowDA on GUDA.}
The active policy network of GFlowDA  is capable of effectively transferring to tasks with varying degrees of label distribution shift and label space mismatch. To prove this, we adjust the degree of label heterogeneity by controlling the JSD distance between the source and target label distribution, denoted as $d_\text{JS}(P_s^Y, P_t^Y)$. As this distance increases, DA tasks become more challenging.
Fig.~\ref{fig:transfer} represents the transferability of our method compared with existing ADA methods in A$\rightarrow$W on Office-31 (More results are available in the appendix).
The term "GFlowDA" refers to directly loading the active policy network pre-trained on the original dataset. "GFlowDA trained" indicates that the policy model has been fine-tuned on the new tasks by training 30 epochs.
Our experimental results demonstrate that GFlowDA can directly transfer to new tasks without training, and outperforms non-learning based methods in most cases. Furthermore, `GFlowDA trained'' improves performance on specific tasks compared to `GFlowDA ''. 
% The transfer of AL strategies is practically challenging, making the advantages of our method even more apparent and highlighting its strong transferability.

\begin{figure*}[htbp]
    \begin{center}
    \includegraphics[width=6.8in, height=1.6in]{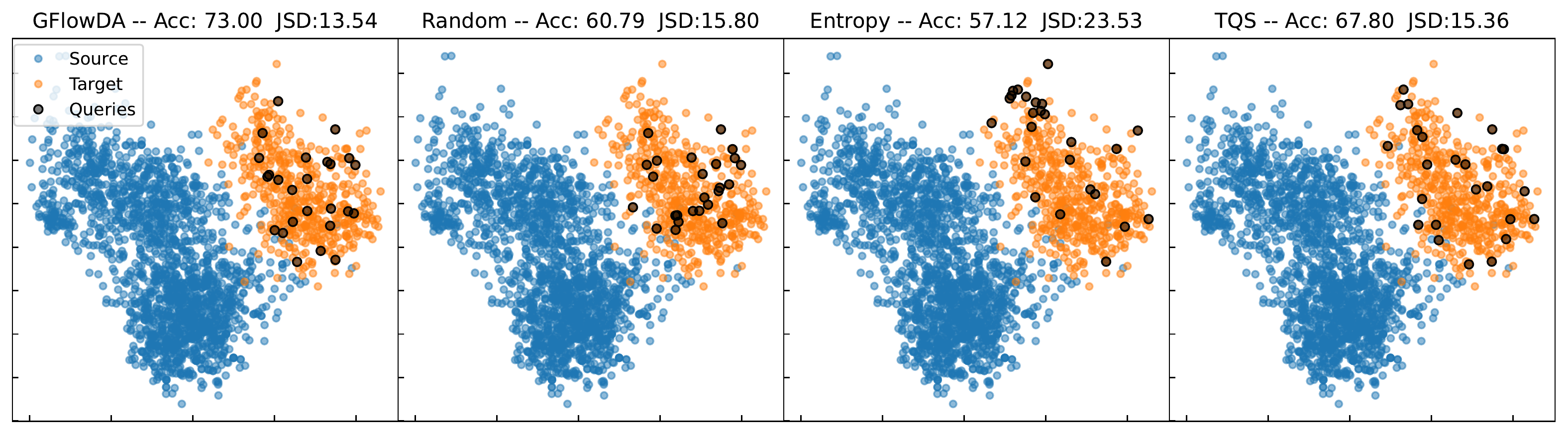}
    \end{center}
    \vspace{-0.2cm}
    \caption{Feature visualization on the Office-31 A $\rightarrow$ W task. Selected samples are shown as black dots in the plots.
    }
    \label{fig:pca}
    \vspace{-0.3cm}
 \end{figure*}

\begin{figure*}[htbp]
    \centering
    \begin{minipage}[t]{0.48\textwidth}
      \centering
      \includegraphics[width=\linewidth]{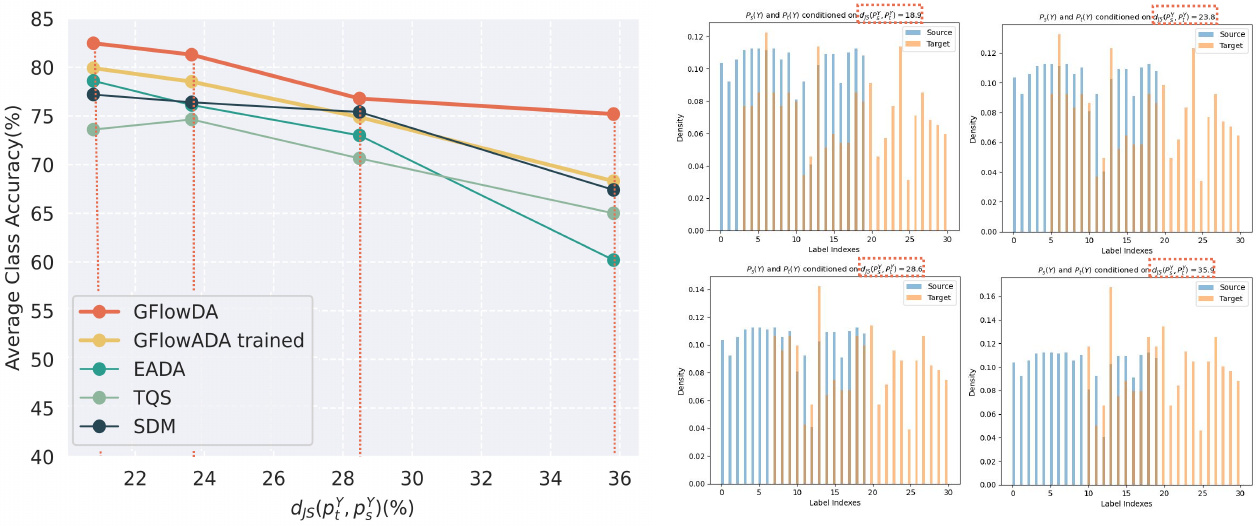}
      \vspace{-0.2cm}
      \caption{\textbf{Left}: Accuracy (\%) with varying $d_\text{JS}(P_s^Y, P_t^Y)$ values. \textbf{Right}: Label distribution with varying $d_\text{JS}(P_s^Y, P_t^Y)$ values.}
      \label{fig:transfer}
    \end{minipage}
    \hfill
    \begin{minipage}[t]{0.48\textwidth}
      \centering
      \includegraphics[width=\linewidth]{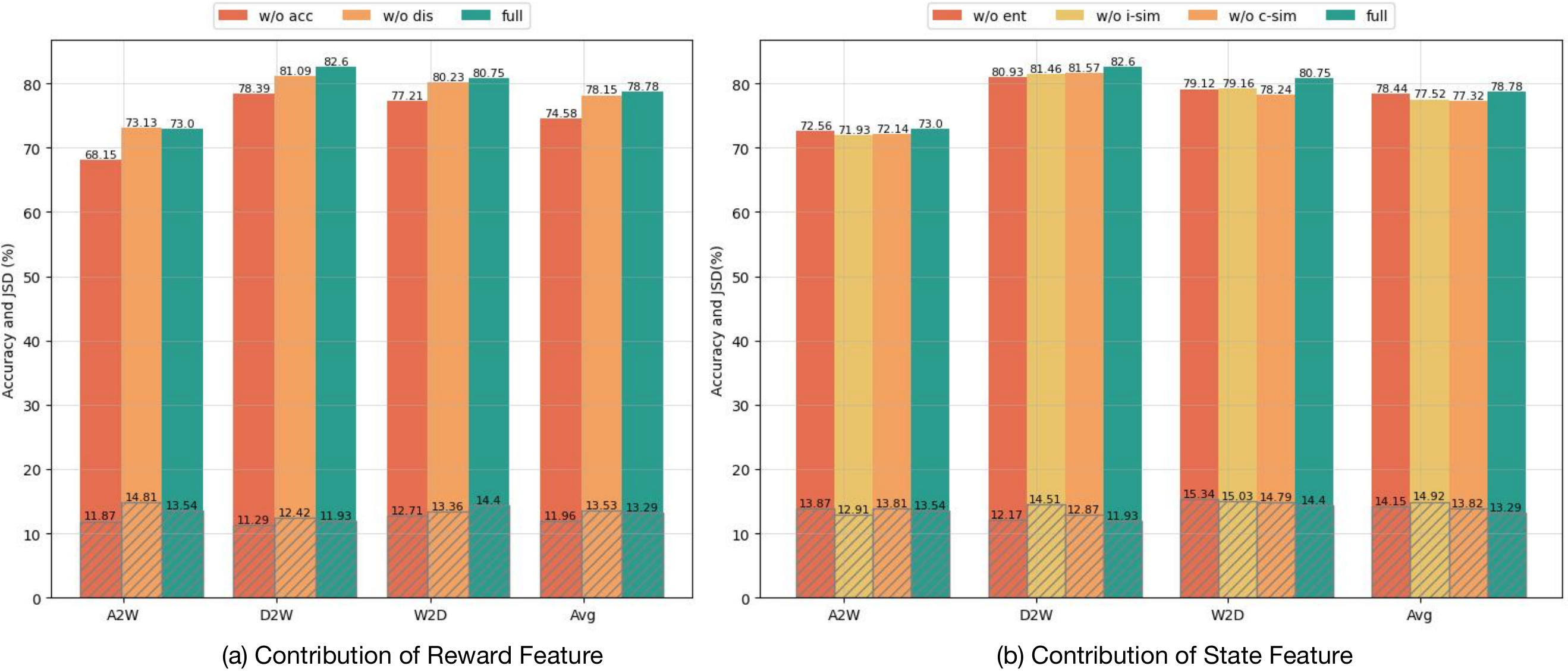}
      \caption{Ablation Study. 
      ``dis'' indicates the MMD distance. ``i-sim'' and ``c-sim'' indicate instance and class-level similarity.
      }
      \vspace{-0.2cm}
      \label{fig:ablation}
    \end{minipage}
    \vspace{-0.3cm}
  \end{figure*}

\subsection{Further Analysis of GFlowDA}

\textbf{Performance of GFlowDA on Existing Settings.}
To demonstrate the effectiveness of GFlowDA in handling the variants included in GUDA,  we evaluated its performance on Office31-Origin and OfficeHome-RUST. The former represents the original setting without label heterogeneity, while the latter represents the GLS scenario~\cite{hwang2022combating}. As shown in Table~\ref{table:exist}, our method outperforms or achieves comparable results to other methods. 
% This highlights the robustness of our approach and its ability to adapt to different domain adaptation settings.

\textbf{Performance of Existing methods on GUDA.}
Due to the absence of AL in previous label space mismatch and label distribution shift methods, as well as differences in prediction space ($|\hat{\mathcal{Y}}| = k+1$), it is unfair and impractical to directly compare GFlowDA with those methods. To ensure fairness and gain a deeper understanding of GFlowDA, we analyze AUDA and LAMDA separately, which are AL methods developed for the UniDA and GLS settings respectively. As shown in the Tbale~\ref{table:exist}, their performance significantly decreases when applied to the GUDA setting. 
% This fully demonstrates that GUDA is a more realistic and
% complex scenario and can guide us in proposing GFlowDA for solving all the variants.

\textbf{Qualitative Analysis}
As illustrated in Fig.~\ref{fig:pca}, we present the feature visualization on Office-31 A $\rightarrow$ W with 5\% budget. It is obvious that the samples selected by GFlowDA can effectively restore the target label distribution, thus contributing to addressing GUDA problems. 
In comparison, the Random method provides a certain level of estimation for the target distribution compared to Entropy, it lacks stability and does not consider sample informativeness. 
The Entropy and TQS methods prioritize prediction uncertainty in  sample selection, resulting in the inclusion of distant samples from the source domain. Consequently, these methods may not fully restore the target label distribution, leading to suboptimal performance.

\textbf{Varying the Label Budget.}
To demonstrate the effectiveness of GFlowDA, we conducted experiments with varying the labeling budget from 0\% to 10\%, as shown in Figure~\ref{fig:budget}. 
Across both Office-31 and Office-Home datasets, GFlowDA consistently outperforms baselines in terms of accuracy and JSD, showcasing GFlowDA can provide excellent performance across varying labeling budgets, making it a promising solution to address GUDA.

% \subsection{Ablation Study of GFlowDA}

\textbf{Contribution of State and Rewad Features.}
We further investigate the impact of the state features introduced in Section ~\ref{sec:framework}.
To study how each feature affects the learned policy, we remove them from the state space individually and examine the resulting performance.
The experimental results are illustrated in Figure ~\ref{fig:ablation} (b), which shows the performance of GFlowDA with three variants of the state on Office-31.
It can be observed that removing any of the features can result in a drop in accuracy.
We further analyzed the contribution of reward on GFlowDA's performance. Figure~\ref{fig:ablation} (a) shows the results.
Ignoring either component would lead to a passive impact on the overall performance of GFlowDA. 
Overall, it is suggested that the unique combination of state and reward can effectively address GUDA.

\section{Conclusion}
\label{sec:conclusion}
In this work, we propose a comprehensive problem GUDA, which aims to obtain accurate predictions for unknown categories while addressing label heterogeneity. We develop an AL domain adaptation method, GFlowDA, by leveraging GFlowNets' exploration capabilities. To achieve this, we propose a design paradigm of state and reward, along with an efficient solution for parent exploration and state transition. Additionally, GFlowDA introduces a training framework named GUAN for GUDA. 
Experimental results demonstrate GFlowDA's superior performance in five benchmarks. 

\section*{Acknowledgments}
This work was supported by the National Key Research and Development Project of China (2021ZD0110505), National Natural Science Foundation of China (U19B2042), the Zhejiang Provincial Key Research and Development Project (2023C01043), University Synergy Innovation Program of Anhui Province (GXXT-2021-004), Academy Of Social Governance Zhejiang University, Fundamental Research Funds for the Central Universities (226-2022-00064, 226-2022-00051, 226-2022-00142).

\clearpage
%%
%% The next two lines define the bibliography style to be used, and
%% the bibliography file.
\bibliographystyle{ACM-Reference-Format}
\balance
\bibliography{sample-base}

%%
%% If your work has an appendix, this is the place to put it.
\appendix
\clearpage

\section{Proof of Theorem 1}
\label{theoretical}

Before we give the proof of Theorem~\ref{theorem1}, we first present the error decomposition theorem proposed by \cite{tachet2020domain} under the assumption that $\mathcal{Y}_s = \mathcal{Y}_t$, which is stated as follows.

\begin{theorem}[Error Decomposition Theorem \cite{tachet2020domain}]\label{theorem2}
     For any classifier $\widehat{Y}=(h \circ g)(X)$,
\[
\begin{split}
&\left|\epsilon_s(h \circ g)-\epsilon_t(h\circ g)\right| \\
&\leq\left\|P_s(Y)-P_t(Y)\right\|_1 \cdot \varepsilon_{s}(\widehat{Y} \| Y)+2(k-1) \Delta_{s,t}(\widehat{Y}\| Y).
\end{split}
\]
% where $\left\|{P}_s(Y)-{P}_t(Y)\right\|_1:=\sum_{i \in \mathcal{Y}}\left|P_s(Y=i)-P_t(Y=i)\right|$ is the $L_1$ distance between $P_s(Y)$ and $P_t(Y)$.
\end{theorem}

\noindent
\textbf{Proof of Theorem~\ref{theorem1}.}  Followed by \cite{tachet2020domain}, we denote $\gamma_{s,j} = P_s(Y=j)$ for simplicity. The following identity holds for $a \in\{s,l, t\}$ by the law of total probability:
\[
\begin{aligned}
\epsilon_a(h \circ g)
&=\mathbb{E}_{({X}, {Y}) \sim P_a} \ell({h} \circ g({X}), {Y}) \\
&=\mathbb{E}_{({X}, {Y}) \sim P_a} P_{a}(\widehat{Y} \neq Y)\\
&=\sum_{i \neq j} P_{a}(\widehat{Y}=i, Y=j) \\&=\sum_{i \neq j} \gamma_{a, j} P_{a}(\widehat{Y}=i \mid Y=j).
\end{aligned}       
\]

We further decompose the target risk $\epsilon_t(h \circ g)$ and the source risk $\epsilon_s(h \circ g)$ based on their common and private label space. 
Then we have:
\[
\begin{aligned}
&\epsilon_t(h \circ g)-\epsilon_s(h\circ g) \\
=& \mathbb{E}_{({X}, {Y}) \sim \mathcal{D}_t} P_t(Y \neq \widehat{Y})-\mathbb{E}_{({X}, {Y}) \sim \mathcal{D}_s}P_s(Y \neq \widehat{Y}) \\
=&\sum_{i \neq j} \gamma_{t, j} P_t(\widehat{Y}=i| Y=j)-\sum_{i \neq j} \gamma_{s, j} P_s(\widehat{Y}=i| Y=j) \\
\leq & \left|\sum_{i \neq j, j\in \mathcal{Y}} \gamma_{t, j} P_t(\widehat{Y}=i|Y=j)-\sum_{i \neq j, j\in \mathcal{Y}} \gamma_{s, j} P_s(\widehat{Y}=i|Y=j)\right| \\
+&\sum_{i \neq j, j\in \overline{\mathcal{Y}}_t} \gamma_{t, j} P_t(\widehat{Y}=i|Y=j) - \sum_{i \neq j, j\in \overline{\mathcal{Y}}_s} \gamma_{s, j} P_s(\widehat{Y}=i|Y=j). \\
\end{aligned}
\]
Based on Theorem \ref{theorem2}, we can obtain:
\[
\begin{aligned}
&\left|\sum_{i \neq j, j\in \mathcal{Y}_c} \gamma_{t, j} P_t(\widehat{Y}=i|Y=j)-\sum_{i \neq j, j\in \mathcal{Y}_c} \gamma_{s, j} P_s(\widehat{Y}=i|Y=j)\right| \\
% \leq& \sum_{i \neq j, j\in \mathcal{Y}}\left|\gamma_{s, j}-\gamma_{t, j}\right| \cdot\left( \alpha_j
% P_s^{ji}+\beta_{j} P_t^{ji}\right)+
% 2(k-1)  \Delta^{\mathcal{Y}}_{{s},t}(\widehat{y}) \\
\leq & \left\|P_s(Y_c)-P_t(Y_c)\right\|_1 \cdot \varepsilon_{s}(\widehat{Y}\|Y_c)+2(k-1) \Delta_{{s}, t}(\widehat{Y}\|Y_c),
\end{aligned}
\]
% where $Y_c \in \mathcal{Y}_c$ indicates a random variable taking value from the common label space $\mathcal{Y}_c$ and $k = |\mathcal{Y}_c|$ is the common label space size.
Using the above inequality, we have:
\[
\begin{aligned}
&\mathbb{E}_{({X}, {Y}) \sim \mathcal{D}_t} P_t(Y \neq \widehat{Y}) - \mathbb{E}_{({X}, {Y}) \sim \mathcal{D}_s}P_s(Y \neq \widehat{Y}) \\
\leq & \left\|P_s(Y_c)-P_t(Y_c)\right\|_1 \cdot \varepsilon_{s}(\widehat{Y}\|Y_c)+2(k-1) \Delta_{{s}, t}(\widehat{Y}\|Y_c)\\
+&\sum_{i \neq j, j\in \overline{\mathcal{Y}}_t} \gamma_{t, j} P_t(\widehat{Y}=i|Y=j) - \sum_{i \neq j, j\in \overline{\mathcal{Y}}_s} \gamma_{s, j} P_s(\widehat{Y}=i|Y=j). \\
\end{aligned}
\]
Recall that the source risk can be decomposed into two parts:
\[
\begin{aligned}
 & \mathbb{E}_{({X}, {Y}) \sim \mathcal{D}_s}P_s(Y \neq \widehat{Y}) = \\
& \sum_{i \neq j, j\in \overline{\mathcal{Y}}_s} \gamma_{s, j} P_s(\widehat{Y}=i|Y=j) +
  \sum_{i \neq j, j\in {\mathcal{Y}}_c} \gamma_{s, j} P_s(\widehat{Y}=i|Y=j).
\end{aligned}
\]
Combining the above inequality and identity, we have:
\[
\begin{aligned}
&\mathbb{E}_{({X}, {Y}) \sim \mathcal{D}_t} P_t(Y \neq \widehat{Y}) \\
\leq & \left\|P_s(Y)-P_t(Y)\right\|_1 \cdot \varepsilon_{s}(\widehat{Y}\|Y)+2(k-1) \Delta_{{s}, t}(\widehat{Y}\|Y) \\
    +& \sum_{i \neq j, j\in \overline{\mathcal{Y}}_t} \gamma_{t, j} P_t(\widehat{Y}=i|Y=j) + \sum_{i \neq j, j\in {\mathcal{Y}_c}} \gamma_{s, j} P_s(\widehat{Y}=i|Y=j).
\end{aligned}
\]
For simplicity, we still use $\epsilon_s(h \circ g)$ to denote the classification error on common label space, i.e., the last term of the above inequality.  We have:
\[
\begin{aligned}
&\epsilon_t(h \circ g) 
\leq   \epsilon_{s}(h \circ g) \\
+&\left\|P_s(Y)-P_t(Y)\right\|_1 \cdot \varepsilon_{s}(\widehat{Y}\|Y)+2(k-1) \Delta_{{s}, t}(\widehat{Y}\|Y) \\
+& \sum_{i \neq j, j\in \overline{\mathcal{Y}}_T} \gamma_{t, j} P_t(\widehat{Y}=i|Y=j).
\end{aligned}
\]
Moreover, we have 
$$\sum_{i \neq j, j\in \overline{\mathcal{Y}}_T} \gamma_{t, j} P_t(\widehat{Y}=i|Y=j) \leq \varepsilon_{{t}}(\widehat{Y}||\overline{Y}_t),
$$
where 
$$
\varepsilon_{t}(\widehat{Y}||\overline{Y}_t):=\max_{j \in \overline{\mathcal{Y}}_t}P_s(\widehat{Y} \neq Y \mid Y=j),
$$
and hence,
\[
\begin{aligned}
&\epsilon_t(h \circ g) 
\leq   \epsilon_{s}(h \circ g) \\
+&\left\|P_s(Y)-P_t(Y)\right\|_1 \cdot \varepsilon_{s}(\widehat{Y}\|Y)+2(k-1) \Delta_{{s}, t}(\widehat{Y}\|Y) \\
+&\varepsilon_{{t}}(\widehat{Y}||\overline{Y}_t).
\end{aligned}
\]
It is worth noting that the above target upper bound only involves the source domain $\mathcal{D}_s$.
Since AL is required in GUDA, an additional domain $\mathcal{D}_l$ corresponding to the selecting labeled target data is introduced.
 Similar to the above inequality, for the error decomposition between the target and the selected labeled target domain, we have:
\[
\begin{aligned}
& \epsilon_t(h \circ g)
\leq   \epsilon_{l}(h \circ g) \\
+&\left\|P_l(Y_l)-P_t(Y_l)\right\|_1 \cdot \varepsilon_{l}(\widehat{Y}\|Y_l)+2(v-1) \Delta_{{l}, t}(\widehat{Y}\|Y_l) \\
+& \varepsilon_{{t}}(\widehat{Y}||\overline{Y}_t).
\end{aligned}
\]
Note that the common label space between the selected target domain $\mathcal{D}_l$ and the original target domain $\mathcal{D}_t$ is $\mathcal{Y}_l$ due to $\mathcal{Y}_l \subseteq \mathcal{Y}_t$.
Finally, combining the above two inequalities, we get:
\begin{align}
& \epsilon_t(h \circ g) 
\leq  \epsilon_{s}(h \circ g)+  \epsilon_{l}(h \circ g) + \delta_{s,t} + \delta_{l,t} +\varepsilon_{{t}}(\widehat{Y}||\overline{Y}_t), \nonumber
\end{align}
where 
\begin{align}
    \delta_{s,t} &= \left\|P_s(Y_c)-P_t(Y_c)\right\|_1 \cdot \varepsilon_{{s}}(\widehat{Y}||Y_c) + 2(k-1) \Delta_{{s}, t}(\widehat{Y}||Y_c)  \nonumber \\
    \delta_{l,t} &= \left\|P_{l}(Y_l)-P_t(Y_l)\right\|_1 \cdot \varepsilon_{{l}}(\widehat{Y}||{Y}_{l}) + 2(v-1) \Delta_{{l}, t}(\widehat{Y}||Y_l). \nonumber
\end{align}
Above all, we complete the proof of Theorem~\ref{theorem1}.

\begin{table*}
        \centering
        \tiny
        \caption{Remaining Results on \textbf{Office-Home} with 5\% budget. \textbf{Boldface} and \underline{underline} represent
        the best and second best scores.}

            \begin{subtable}[t]{\linewidth}
            \resizebox{17.5cm}{!}{
                \begin{tabular}{lcccccccccccccc}
                \toprule
                \multirow{2}{*}{Method} &
        % \multicolumn{12}{c}{\textbf{Office-Home}}  
        % % \multicolumn{2}{c}{PACS}
        % \\
        % \cmidrule{2-13}
        
        %  &
          \multicolumn{2}{c}{Pr$\rightarrow$ Ar} &
          \multicolumn{2}{c}{Pr$\rightarrow$ Cl} &
          \multicolumn{2}{c}{Pr$\rightarrow$ Rw} &
          \multicolumn{2}{c}{Rw$\rightarrow$ Ar} &
          \multicolumn{2}{c}{Rw$\rightarrow$ Cl} &
          \multicolumn{2}{c}{Rw$\rightarrow$ Pr} &
          \multicolumn{2}{c}{Avg} 
        %   &\multicolumn{2}{c}{P$\rightarrow$ A} 
          \\  
        
        % \cmidrule(r){2-3}\cmidrule(r){4-5}\cmidrule(r){6-7}\cmidrule(r){8-9}\cmidrule(r){10-11}\cmidrule(r){12-13}
        
        &Acc        &JSD &Acc        &JSD        &Acc        &JSD         &Acc        &JSD         &Acc        &JSD        &Acc        &JSD        &Acc        &JSD           \\ 
        \midrule
        {Random}      & \underline{50.23}&12.42 &  45.80  & 7.08    &   51.25   & 7.18   &    40.39  & 11.01 &   41.91 & 6.23     &  60.89   &     5.94 &   52.20 &  7.76                     \\
        
        % BvSB       &      &     &      &    &      &  &    &      &     &      &    &      &                    \\
        % 
        {Entropy~\cite{wang2014flexible}}        &36.19 & 14.89 &  40.56   &  15.91   &   57.29   & 18.16   &   37.46   & 17.79 &   41.09 &  15.44    &  56.73   & 18.22    & 48.36   &   14.72                     \\
        {TQS~\cite{fu2020learning}}  & 45.03 & 11.81 &  45.56    & 9.31     &  58.16   &  6.78    & 42.73 & 11.42 & 43.41  &  5.75    &  64.97  & 6.23     &   52.49  & 8.04               \\
        {CLUE~\cite{prabhu2021active}}   & 48.14 & \underline{9.54} &  46.74    &  8.17   &  62.12    &  5.73  &   42.78   & 12.48 &  42.81  &  5.19    &  66.91   &   5.11   &   53.98 &  \underline{7.16}                                   \\
        {EADA~\cite{xie2022active}}  & 42.15 & 11.52 &  40.22   &   7.53  &    55.94  &  14.28  &    42.15  & 16.88 &   36.13 &    5.60  &   64.85  &  8.95    & 50.28 & 9.32                  \\
        % \texttt{DBAL}    &    44.13  &  10.12   &   58.74   &  12.34   &  40.77    &  11.42    &  41.07    &6.44  &  64.46  &   7.31   &  51.66   & 8.88     &              \\            
        {SDM-AG~\cite{xie2022active}}  & 42.84 & 10.83 &  40.92   &  8.81   & 55.81  &  12.12  &   40.65   &  10.42 &  39.25   &  7.33    &  60.69   &  11.92    & 42.84   &10.83                                              \\
        % AUDA   &      &     &      &    &      &  &    &      &     &      &    &      &                   \\
                            
        {LAMDA~\cite{hwang2022combating}}   & 43.75&10.88 & 46.78   &   8.75 &   63.46   &  16.31  &  49.12    & 17.18 &  38.47  & 5.04     &    67.79 &    10.73  &  53.86  &  11.37                  \\   
        \midrule
        % {RLADA(Ours)} &  45.23  &  6.52    &  60.39   &  7.28    & 47.73   &  9.22    & 43.89 &  6.17  &     65.28 &  6.09   &   55.32   &   7.68 \\ 
        % {GFlowDA(Ours)}  &  50.36    & 5.29    &  65.14 &  4.69  & 52.03     & 7.52 &  50.79  &  4.33  &   69.46 &  3.28    &   61.40 &  5.65          \\

        {RLADA (Ours)} & 47.92& 9.87&  45.23  &  \underline{6.52}    &  60.39   &  7.28    & 47.73   &  \underline{9.22}    & \underline{43.89} &  6.17  &     65.28 &  6.09   &   \underline{55.32}   &   7.68   \\ 
        {GFlowDA (Ours)}  & \textbf{50.78}&\textbf{8.78} &  \textbf{48.36}    & \textbf{5.29}    &  \textbf{63.14} &  \textbf{4.69}  & \textbf{50.03}     & \textbf{7.52} &  \textbf{48.79}  &  \textbf{4.33}  &   \textbf{67.46} &  \textbf{3.28}    &   \textbf{59.32} &  \textbf{5.65}          \\
        
        \bottomrule
        
      \end{tabular}}
            \end{subtable}
            \label{tab:table_all_oh}
\end{table*}

\section{Experiments}
\label{sec:experiment}

\subsection{Dataset Setup under GUDA}
\label{dataset}

\textbf{Office-31:}  For Office-31, we use the middle 10 classes as the common label set, i.e., $\mathcal{Y} = \{10-19\}$, then in alphabetical order, the first 10 classes are used as the source private classes, i.e., $\overline{\mathcal{Y}}_s = \{0-9\}$, and the rest 11 classes are used as the target private classes, i.e., $\overline{\mathcal{Y}}_t = \{20-31\}$. Then we only consider 30\% of source samples in common classes \{0-4\} and private classes \{10-14\}.

\begin{table*}
    \centering
    \caption{Remaining Results on PACS with 1\% labeling budget. \textbf{Boldface} and \underline{underline} represent
    the best and second best scores.}

        \begin{subtable}[t]{\linewidth}

        \centering
        \resizebox{17.5cm}{!}{
            \tiny
            \begin{tabular}{lccccccccccccccccccc}
            \toprule
            \multirow{2}{*}{Method} &
    % \multicolumn{12}{c}{\textbf{PACS}} 
    % &  \multicolumn{2}{c}{\textbf{VisDA}}
    % \\
    % \cmidrule{2-13}
    
    %  &
    \multicolumn{2}{c}{C $\rightarrow$ S} &
    \multicolumn{2}{c}{P $\rightarrow$ A} &
      \multicolumn{2}{c}{P $\rightarrow$ C} &
      \multicolumn{2}{c}{P $\rightarrow$ S} &
      \multicolumn{2}{c}{S $\rightarrow$ A} &
      \multicolumn{2}{c}{S $\rightarrow$ C} &
      \multicolumn{2}{c}{S$\rightarrow$  P} &
      \multicolumn{2}{c}{Avg} \\  
    
    % \cmidrule(r){2-3}\cmidrule(r){4-5}\cmidrule(r){6-7}\cmidrule(r){8-9}\cmidrule(r){10-11}\cmidrule(r){12-13}\cmidrule(r){14-15} 
    
    &Acc        &JSD        &Acc        &JSD  &Acc        &JSD        &Acc        &JSD         &Acc        &JSD         &Acc        &JSD        &Acc        &JSD        &Acc        &JSD         \\ 
    \midrule
    \texttt{Random}     &  50.14 & \textbf{0.44}     &   54.46   & 7.46     &    60.36  & 6.62    &  71.87    &3.05    &   49.65   & \underline{2.52} &  53.81  & 4.84     &  82.88   &  \underline{3.10}    & 63.27   &   4.30           \\
    
    % BvSB       &      &     &      &    &      &  &    &      &     &      &    &      &                    \\
    % 
    \texttt{Entropy~\cite{wang2014new}}      &  49.98  & 6.21&  53.89   &2.33        &  \underline{63.82}   &  6.78   &   \textbf{79.98}   &2.24    &  45.76    & 3.40 & 50.07   & 16.55     & 64.74    &  17.86    & 58.38   & 8.20                    \\
    \texttt{TQS~\cite{fu2021transferable}}   & \textbf{74.76} &   2.27   & 58.77 & 11.08        &   63.68  & 9.23    &   75.34   & 2.38   &   31.19   & 10.57 &   36.45 &  9.67    &  56.92   &    9.05  &   63.12 & 6.37         \\
    % \texttt{CLUE~\cite{prabhu2021active}}   &      &     &      &    &      &  &    &      &     &      &    &      &                                            \\
    \texttt{EADA~\cite{xie2022active}}   &33.10   &    5.80 &  27.09  & 7.17      &  24.77  &  7.69   &   20.98   & 5.04   &  46.04    &5.51  & 40.25   &  5.31    &   53.87  & 14.01     &  34.38  &  11.16             \\
                
    \texttt{SDM-AG~\cite{xie2022learning}}  &  \underline{69.88}  &  \underline{2.10}    &  41.08 &7.08  & 58.46    &  4.49   &  74.84    & 4.31   &   43.00   &  \textbf{1.74}& 46.53   & \underline{1.05}    &   54.18  & 6.04     &63.16    & 4.18                                           \\
    % AUDA   &      &     &      &    &      &  &    &      &     &      &    &      &                   \\
                        
    \texttt{LAMDA~\cite{hwang2022combating}}   & 46.37 & 2.31 &\underline{60.21}   &    3.05    &  59.01   &   \underline{3.48}  &    \underline{76.73}  & 6.29   &   \underline{62.41}   &9.25  &  55.87  &   5.93   &   \underline{89.40}  &  17.77    &  65.97  & 5.85  \\   
    \midrule
    \texttt{RLADA (Ours)}  &   44.52 &  2.26    & 59.21 & \underline{2.18}     &58.41    &   3.81  &     58.57 &  \underline{2.16}  &  62.20    & 2.85 &  \underline{59.32}  &  1.41    &  88.57& 5.32 &  \underline{66.00}  & \underline{2.94}        \\ 
    \texttt{GFlowDA (Ours)}  &45.61   & 3.51   &    \textbf{62.36}  &  \textbf{0.74}        &  \textbf{64.03}  &   \textbf{2.92}  &    59.03  &   \textbf{1.99} &   \textbf{63.15}   & 3.12 &  \textbf{61.82}  &   \textbf{1.03}   &  \textbf{93.02}   &    \textbf{1.66}  &  \textbf{68.26}  & \textbf{1.93}              \\
    
    \bottomrule
            \end{tabular}}
        \end{subtable}
        \label{tab:table_all_pacs}
\end{table*}

\textbf{Office-Home:} For Office-Home, we use the middle 10 classes as the common classes, i.e., $\mathcal{Y} = \{30-39\}$. Then the first 30 classes are used as the source private classes, i.e., $\overline{\mathcal{Y}}_s = \{0-29\}$. Correspondingly, the last 25 classes are used as the target private classes, i.e., $\overline{\mathcal{Y}}_t = \{40-64\}$. To construct a large label distribution, we consider 30\% of source samples in common classes \{0-14\} and private classes \{30-34\}, 30\% of target samples in common classes \{35-39\} and private classes \{40-49\}.

\textbf{PACS:} For PACS, we only consider one class labeled as ``2'' as the common label space to construct a large label space gap, which means that $\mathcal{Y} = \{2\}$. We use the first two classes as the source private classes denoted by $\overline{\mathcal{Y}}_s = \{0,1\}$ and use the last four classes as the target private classes denoted by $\overline{\mathcal{Y}}_t = \{3,4,5,6\}$. We consider 30\% of source samples in class 0 and 30\% of target samples in common class 2 and private class 3.

\textbf{VisDA:} For VisDA, we use classes 5 and 6 as the common classes denoted by $\overline{\mathcal{Y}}_s = \{5,6\}$. The first five classes are used as the source private classes denoted by $\overline{\mathcal{Y}}_s = \{0-4\}$. The last five classes are used as the target private classes denoted by $\overline{\mathcal{Y}}_t = \{7-11\}$. We consider 30\% of source samples in common class 5 and source private classes \{0,1,2\} and consider 30\% of target samples in common class 5 and target private classes \{7,8,9\}.

\subsection{Comparison Baselines} 
\label{comp}
We compare GFlowDA against several AL and ADA methods including
(1) Random, (2) Entropy~\cite{wang2014new}, (3) TQS~\cite{fu2021transferable}, (4) CLUE~\cite{prabhu2021active}, (5) EADA~\cite{xie2022active}, (6) SDM-AG~\cite{xie2022learning}, (7) LAMDA~\cite{hwang2022combating}. 
Furthermore, to illustrate the superiority of GFlowDA, we also implemented a new algorithm by changing the policy network from GflowNet to Proximal Policy Optimization~\cite{schulman2017proximal} and keeping everything else the same, named (8) Reinforcement Learning Active Domain Adaptation, abbreviated as RLADA. RLADA is proposed by us to fill the gap of learning-based approaches in the ADA literature. We compare GFlowADA with RLADA to reflect the advantages of GFlowNet over traditional reinforcement learning algorithms such as Proximal Policy Optimization(PPO)~\cite{schulman2017proximal}.

\subsection{Implementation Details}
\label{implem}

\textbf{Domain Adaptation Model:}
For Random, Entropy, LAMDA and GFlowDA, we apply ResNet50~\cite{he2016deep} models pre-trained on ImageNet ~\cite{krizhevsky2012imagenet} as a feature extractor. We use Adadelta optimizer training with a learning rate of 0.1 and  a batch size of 32.  The classifier is implemented by a fully-connected layer. The domain discriminator contains  a fully-connected layer and a sigmoid activation layer. 
For other active domain adaptation methods, we use default hyperparameters and network architecture introduced in their works except that keeping using ResNet50 pretrained on ImageNet as the backbone.
We train 1 epoch in the training process for GFlowDA and 40 epochs for other baselines.

\textbf{Policy Network Model:}
For GFlowDA and RLADA, we implement the policy network as a two-layer MLP with a hidden layer size of 8. We use Adam as the optimizer with a learning rate of 0.001. The policy network is trained for a maximum of 2000 episodes with a trajectory size of 5. 
All methods are implemented based on PyTorch, employing ResNet50~\cite{he2016deep} models pre-trained on ImageNet ~\cite{krizhevsky2012imagenet}.
Besides, we run each experiment three times and report mean accuracies and JSD values.
% Due to space limitations, more details about our experimental setup are described in the appendix.

% See appendix for more details about the experimental setup.

\subsection{More Results}

\textbf{Performance of GFlowDA.}
% For the large-scale dataset VisDA, traditional active learning strategies fail to select informative samples to annotate. GFlowDA still achieves the best accuracy than other baselines.
Table \ref{tab:table_all_oh} and Table \ref{tab:table_all_pacs} present the complete experimental results of GFlowDA and compared baselines on all subtasks of the Office-Home and PACS datasets. 
% Table \ref{tab:domainnet} shows the results of GFlowDA and three main SOTA methods on three subtasks of the DomainNet dataset. 
These tables demonstrate the superiority and robustness of GFlowDA, which outperforms other methods in all datasets.

\end{document}